\documentclass{IEEEtran}
\usepackage{amssymb,graphicx,color,booktabs,cite,multirow,multicol,amsfonts,textcomp,changepage,epstopdf,psfrag,algorithm,algorithmic,diagbox,mdwmath,bm,enumerate,colortbl,ragged2e,xcolor,amsmath,times,tikz,mathptmx}

\usepackage[hidelinks]{hyperref}

\usepackage[tight,footnotesize,center]{subfigure}
\definecolor{lime}{HTML}{A6CE39}
\DeclareRobustCommand{\orcidicon}{%
	\begin{tikzpicture}
	\draw[lime, fill=lime] (0,0) 
	circle [radius=0.16] 
	node[white] {{\fontfamily{qag}\selectfont \tiny ID}};
	\draw[white, fill=white] (-0.0625,0.095) 
	circle [radius=0.007];
	\end{tikzpicture}
	\hspace{-2mm}
}

\foreach \x in {A, ..., Z}{
	\expandafter\xdef\csname orcid\x\endcsname{\noexpand\href{https://orcid.org/\csname orcidauthor\x\endcsname}{\noexpand\orcidicon}}
}

\begin{document}
\title{ASPCNet: A Deep Adaptive Spatial Pattern Capsule Network for Hyperspectral Image Classification}
\author{Jinping Wang\orcidA{},~\IEEEmembership{Student Member,~IEEE,} Xiaojun Tan\orcidB{}, \\ Jianhuang Lai\orcidD{},~\IEEEmembership{Senior Member,~IEEE,} Jun Li\orcidE{}, and Canqun Xiang
\thanks{This work was supported by the Key-Area Research and Development Program of Guangdong Province under Grant 2020B090921003 and Southern Marine Science and Engineering Guangdong Laboratory (Zhuhai) under Grant SML2020SP011. (Corresponding author: Jun Li.)}

\thanks{J. Wang, X. Tan, and J. Li are with the School of Intelligent Systems Engineering, Sun Yat-sen University, Guangzhou, 510006, China (e-mail: wangjp29@mail2.sysu.edu.cn; tanxj@mail.sysu.edu.cn; stslijun@mail.sysu.edu.cn).\par
J. Lai is with the School of Computer Science and Engineering, Sun Yat-sen University, and also with the Key Laboratory of Machine Intelligence and Advanced Computing, Ministry of Education, Sun Yat-sen University, Guangzhou, 510006, China (e-mail: stsljh@mail.sysu.edu.cn).\par
J. Wang and X. Tan are with Southern Marine Science and Engineering Guangdong Laboratory (Zhuhai), Zhuhai, 519082, China.\par
C. Xiang is with the College of Electronics and Information Engineering, Shenzhen University, Shenzhen 518060, China (e-mail: xiangcanqun2018@email.szu.edu.cn)}
}

\markboth{ ~Vol.~X, No.~X, 2021}
{Shell \MakeLowercase{\textit{et al.}}:}

\maketitle

\begin{abstract}
  Previous studies have shown the great potential of capsule networks for the spatial contextual feature extraction from {hyperspectral images (HSIs)}.  However, the sampling locations of the convolutional kernels of capsules are fixed and cannot be adaptively changed according to the inconsistent semantic information of HSIs. Based on this observation, this paper proposes an adaptive spatial pattern capsule network (ASPCNet) architecture by developing an adaptive spatial pattern (ASP) unit, that can rotate the sampling location of convolutional kernels on the basis of an enlarged receptive field. Note that this unit can learn more discriminative representations of HSIs with fewer parameters. Specifically, two cascaded ASP-based convolution operations (ASPConvs) are applied to input images to learn relatively high-level semantic features, transmitting hierarchical structures among capsules more accurately than the use of the most fundamental features. Furthermore, the semantic features are fed into ASP-based conv-capsule operations (ASPCaps) to explore the shapes of objects among the capsules in an adaptive manner, further exploring the potential of capsule networks. Finally, the class labels of image patches centered on test samples can be determined according to the fully connected capsule layer. Experiments on three public datasets demonstrate that ASPCNet can yield competitive performance with higher accuracies than state-of-the-art methods. 
	\end{abstract}


\section{Introduction}
 \IEEEPARstart{H}{yperspectral} remote sensing technology has attracted much attention in recent years since it can include hundreds of contiguous spectral bands and capture more accurate and discriminative features for different objects, especially compared with panchromatic and multi-spectral images. Hyperspectral image classification (HSIC), which refers to automatically assigning a specific label for each pixel in a scene, has become an active topic in many research fields, such as defense and security \cite{du2016beyond}, intelligent transportation, intelligent healthcare \cite{qureshi2019hyperspectral,han2020hyperspectral}, and forest and unmanned aerial vehicle monitoring (e.g., mangrove biomass estimation) \cite{wang2019review,liu2019mapping}.

 During the last few decades, a large number of pixelwise-based classifiers, mainly based on  spectral signatures, have been proposed for classification tasks, such as support vector machines (SVMs) \cite{svm2004},  logistic regression \cite{li2011spectral}, $k$-nearest-neighbors \cite{spectral-review,DL_review2019}, and random forests \cite{tan2020estimation,RF2013}. {With the improvement in sensor spatial resolution, many spectral-spatial-based classification methods play roles \cite{Morphological,Markov2014,CCJSR,Dictionary,7880591,cao2019hyperspectral,he2017recent}, taking the spatial information of images into consideration and improving the classification accuracies.} As one of the crucial steps in spectral-spatial classification, feature extraction-based methods have attracted much attention. {Some state-of-the-art feature extraction methods, which are mainly based on general machine learning techniques, have achieved good classification performance, such as the image fusion and recursive filtering \cite{IFRF} method, the extended morphological profiles (EMP) \cite{EMP} method, the edge-preserving filter (EPF) \cite{EPF} method, and the superpixel segmentation method\cite{tu2020discriminant,KNNRS}.} However, the aforementioned feature extraction approaches mainly classify an image in a shallow manner, e.g., the feature extractors and classifiers only focus on a single layer. Comparatively, deep learning-based techniques can allow computers to learn different image features through a series of hierarchical layers, and the learning process is totally automatic and efficient. Currently, deep network models are widely used in many image processing tasks \cite{pmlr-v22-bordes12,Girshick_2014_CVPR,Redmon_2016_CVPR,Jiayi2020,Szegedy_2015_CVPR}. 
 Typical deep learning networks for HSIC can be summarized into five categories \cite{SONGWEIWEI}: stacked autoencoders \cite{SAE,SAE2}, deep belief networks\cite{DBN}, recurrent neural networks\cite{RNN2017}, generative adversarial networks\cite{GAN}, and convolutional neural networks (CNNs) \cite{DEEPERCNN}. 

{Unlike other deep learning methods, CNN models possess local receptive fields and shared weight architecture, and have shown strong abilities in the feature extraction process.}  Generally, CNNs can be grouped into two classes: 
(1) The first class extracts spectral-spatial characteristics by simultaneously using 3D filtering. As the layers go deeper, the model features become more precise and reliable \cite{DFFN_song,CNN2015}. { Considering that the pooling operation of CNNs may lose the spatial information of hyperspectral images (HSIs),} the dilated neural networks were introduced for HSIC \cite{fang2019hyperspectral,8950204,devaram2019hyperspectral}, and the core idea of which is to avoid resolution reduction in the pooling layer while enlarging the receptive field through a dilated convolution strategy. Moreover, a multi-scale dilated residual CNN has been proposed to improve further classification performance \cite{8921284}.  (2) The second class extracts the spectral and spatial characteristics and subsequently fuses them. In the literature \cite{bigdata}, a spectral-spatial fully convolutional network is proposed that introduces a dense conditional random field into two parallel flows to balance the obtained spectral and spatial features. In contrast, Li {\em et al.} used a fusion scheme with layer-specific regularization and smooth normalization to adaptively learn the fusion weights of local and global information \cite{8920212}. Moreover, the fusion of multilevel and multiscale-based spatial-spectral CNNs also has  potential for HSI \cite{Yen-Wei-Chen,mu2020multi}.

{Although general CNNs are capable of image feature extraction, some shortcomings still exist that strongly threaten the performance of deep learning methods, such as problems related to the loss of local information caused by the pooling layer and an inability to understand spatial positional relationships among features.} 
To overcome these issues, Sabour {\em et al.} proposed a capsule network (CapsuleNet) \cite{Capsulenet_Sabour} that can achieve results superior to those of CNNs. Later, capsulenet was first introduced for HSIC in \cite{paoletti2018capsule}. However, the original capsulenet includes only one convolution layer and one fully connected capsule layer. {To deepen the method and allow it to solve problems due to the stacking of multicapsule layers, many capsulenet-based expansions have been proposed that can achieve better classification performance \cite{2019deepcapsrajasegaran,arun2019capsulenet,zhu2019deep,zhang2019remote,xu2020faster}.} Motivated by the core ideas of CNNs, a modification of the traditional capsulenet named conv-capsule has been proposed for classification\cite{zhu2019deep,zhang2019remote}, which uses a conv-capsule unit to achieve a local connection and shared transform matrices. Experimental results indicate that the conv-capsule unit provides results competitive with those of traditional capsulenet. Moreover, Xu {\em et al.} designed multiple kernels with parallel convolution to extract image features in a multi-scale way, reducing the redundancy of parameters and achieving high accuracies for HSIC \cite{xu2020faster}. These works have demonstrated the excellent performance of capsulenet.

However, there are still some shortcomings in the existing capsulenets. First, the semantic features fed into primarycaps are extracted by a simple convolutional layer, which may not be efficient in representing and transferring hierarchical structures among capsules, compared with relatively high-level semantic features. Second, although the 3D conv-capsulenet uses local connections and shared weights to fully use the spectral-spatial information, the window size is fixed and {cannot} be adjusted according to the different shapes of input images. Third, the traditional neural networks use pooling layers to enlarge the receptive field while reducing the model complexity. However, some detailed features might disappear after several pooling operations. {To overcome these defects, a novel adaptive spatial pattern capsule network (ASPCNet) has been proposed that consists of the following major steps:\par

First,  relatively high-level features obtained by adaptive spatial pattern convolution operations (ASPConvs) are fed into primarycaps. Next, adaptive spatial pattern-based conv-capsule operations (ASPCaps) are proposed during the dynamic routing process,  adaptively adjusting the convolutional kernel sampling locations. Finally, the class label of pixels can be determined according to the fully connected capsule layer.} The contributions of this paper are described as follows:\par
 \begin{figure}
	\centering
    \includegraphics[scale=0.33]{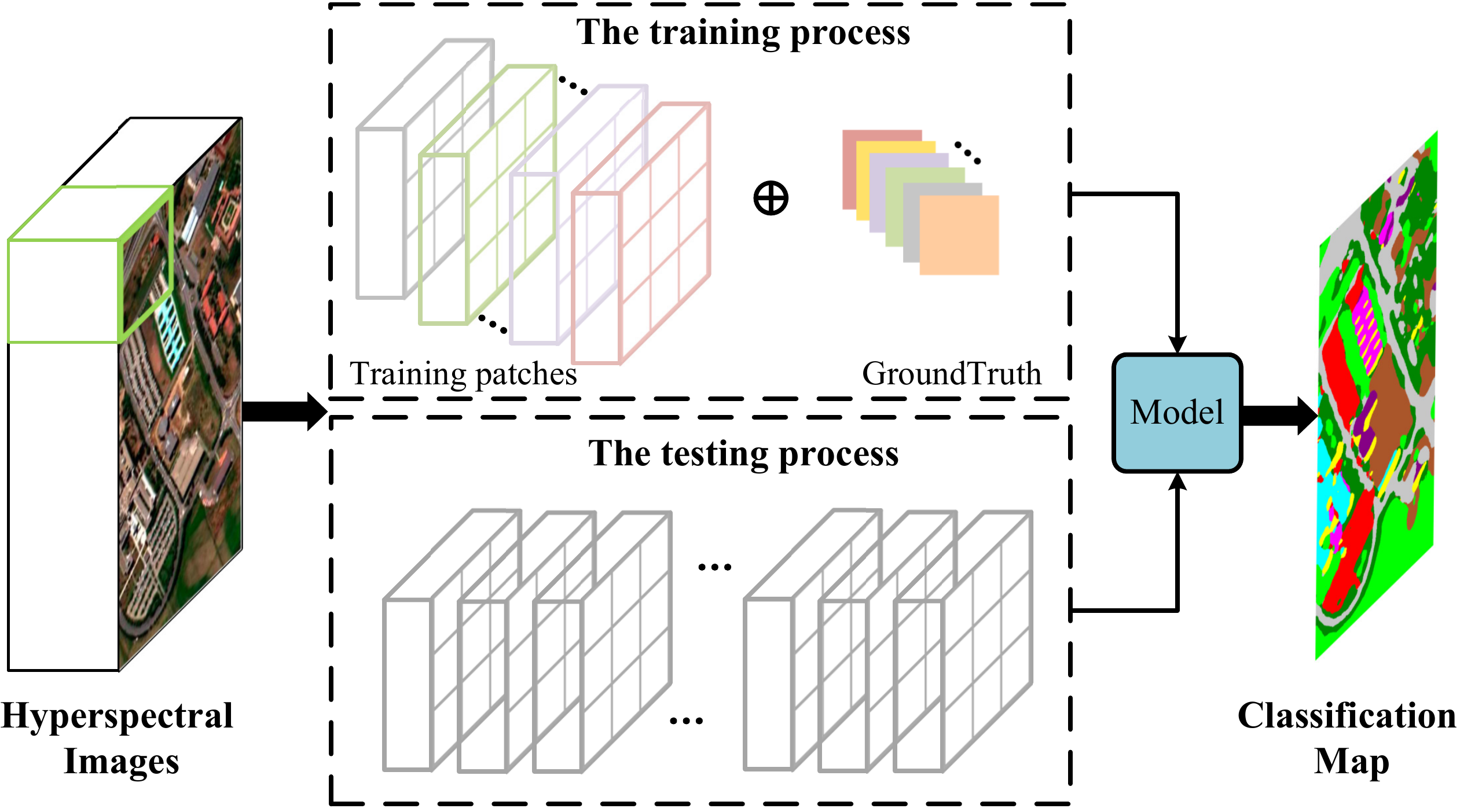}\par
	\caption{The architecture of the supervised learning-based classification of hyperspectral images.}
	\label{fig:coarse_framework}
\end{figure}
1) A novel adaptive spatial pattern (ASP) unit is constructed by adaptively  rotating the sampling location on the basis of dilated convolutions. Instead of using the downsampling strategy, the cascaded ASP layers can simultaneously expand the receptive field and avoid checkerboard effects.

2) A series of ASPConvs are used for feature extraction before the primarycaps layer. The obtained relatively high-level semantic features can offer better services for hierarchical relationship representation among capsules regarding the transmission efficiency and representation ability.\par

3) ASPCaps is developed into a dynamic routing process that can adaptively adjust the shape of local connections according to HSI's complex spatial contexts to ensure that the model has a strong generalization ability.\par

4) Experiments conducted on several real hyperspectral datasets demonstrate that the proposed ASPCNet can always obtain better results and more accurate classification maps than state-of-the-art methods.\par

The remainder of this paper is organized as follows. Section II describes the related works. Section III presents the details of the proposed ASPCNet approach. In Section IV, the experimental results and discussions are provided. Finally, conclusions are given in Section V.

\section{Related Works}
\label{related work}
 \subsection{CNN}
  {CNNs, typical supervised learning methods, use image patches centered on labeled pixels as the training patches for a model training and then the well-trained model can be used for HSIC; see Fig. \ref{fig:coarse_framework}.  Unlike other tools that learn manually designed features, a CNN classifier can learn different types of natural features through a stack of layers \cite{li2016hyperspectral,li2018single,8803839} that includes convolutional layers, pooling layers, and fully connected layers.} In a CNN model, there are a series of fixed boxes named convolution kernels that share the weights of the network to simulate 3-D convolutional operations, and these replicated weights can significantly reduce the number of parameters in its network. The different convolution kernels slide over an input image and can transform this image into another feature domain for effectively modeling these visual features.

 Specifically, we assume that the input image patches (centered on labeled pixels) are of size $N \times N \times C$. In the convolution layers, $k_1$ different filter kernels with windows sizes of $n \times n$ spatially slide over the image patches, obtaining a 3-D tensor of feature maps of size $M \times M \times k_1$. In addition to multiple convolutions, considering the increasing  computational burden, the CNN-based model subsequently employs pooling layers to conduct feature downsampling operations for learning the hierarchy of those different features. After convolutional and pooling layers are achieved, the fully connected layers stack the output values of all previous layers into an $n$-dimensional vector. After a series of fully connected layers, the softmax layer is used to generate the distribution of the probability that the pixel belongs to each class.

\subsection{DCNN}
CNNs are successful models that can effectively extract the detailed features of input images and assign a specific class label for each particular pixel. Suppose the receptive fields of a traditional convolution are of size $N \times N$, { and given a pixel $a$, suppose the image location of $a$ is ($x,y$), and its corresponding value is $p(x,y)$. The output form of centered pixel $a$ can be obtained as follows}
\begin{equation}
O(x,y) = \sum\limits_{({x_n},{y_n}) \in s} {\omega ({x_n},{y_n})}  \cdot p(x + {x_n},y + {y_n}),
\end{equation}
where $s=(x, y) \mid(0,0),(0,1), \ldots,(N-1, N-1)$, which enumerates the location of the kernel and $\omega $ represents the weight of the kernel.\par
\begin{figure}
	\centering
	{\includegraphics[scale=0.35]{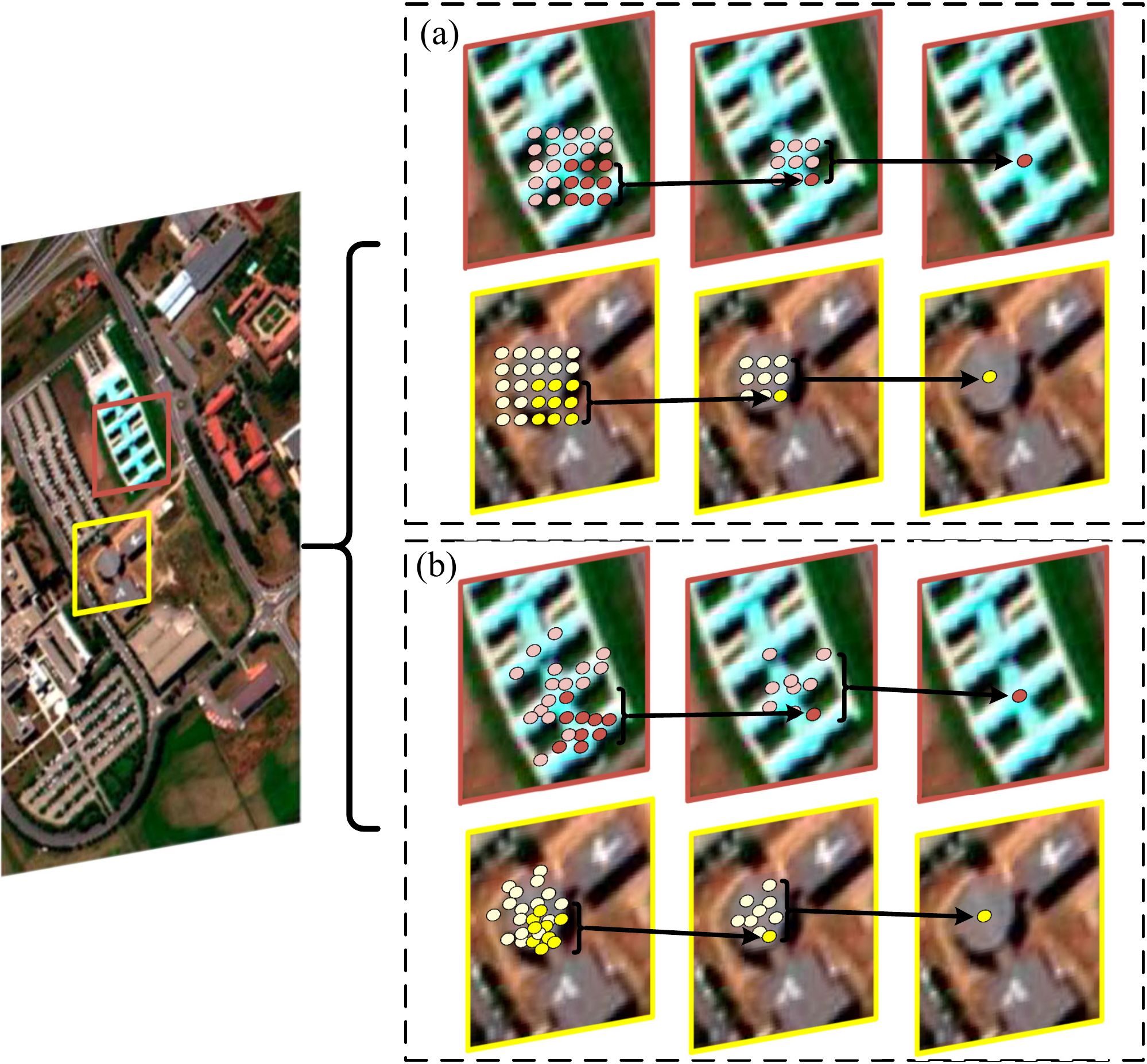}\par
	\caption{Illustration of the differences between different sampling locations on the University of Pavia dataset. (a) Sampling locations based on regular convolution and (b) Sampling locations based on deformable convolution.}
	\label{fig:deformable_HSI}}
\end{figure}
{ However, considering that the spatial shapes and locations of convolutional operations $N \times N$ are fixed, which is not suitable for representing complex image features, deformable convolution neural networks (DCNNs) were conceived in the computer vision field to overcome the problem \cite{Dai_2017_ICCV,Deformable-v2}. Later, the ideal of deformable neural networks was introduced into HSIC\cite{Deformable}, which can effectively extract HSI features, especially for its complex spatial structures.} As we can see from Fig. \ref{fig:deformable_HSI}, there is a contrast between the sampling locations of regular convolution and sampling locations of deformable convolution, in which the deformable convolution is present in the shapes of a rhombus and a circle, showing better coverage. Specifically, two values $\Delta x$, and $\Delta y$  are introduced into the offset field. In this way, the output form of centered $a$ can be changed as follows:
\begin{equation}
O(x,y) = \sum\limits_{({x_n},{y_n} \in s)} {\omega ({x_n},{y_n})}  \cdot p(x + {x_n}+ \Delta  x,y + {y_n}+ \Delta y).
\end{equation}

{However, $\Delta  x$ and $\Delta  y$ are fractional locations, and they cannot directly obtain the real locations. Thus, to avoid checkerboard artifacts, bilinear interpolation is used to obtain these values. The value of $p(\Delta  x, \Delta  y)$ is calculated according to the values of the four surrounding integer locations. The weights of the convolutional filters for generating offset fields are trained based on spatial features to enable the sampling locations to be transferred to similar neighboring pixels. }

\begin{figure*}[htbp]
	\centering{
	{\includegraphics[scale=0.41]{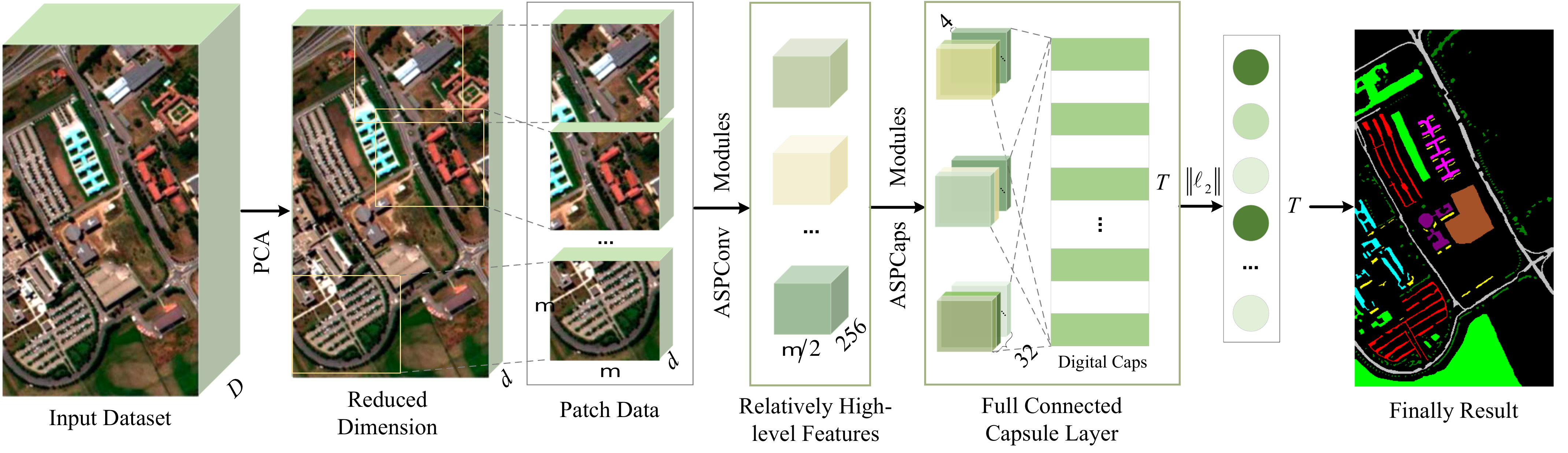}\par
	\caption{The framework of the proposed ASPCNet for HSIC.}}
	\label{fig:framework}}
\end{figure*}

\subsection{CapsuleNet}
 In capsulenets,  specific types of functional neurons are grouped together to form capsules, and individual neurons represent one of various attributes of a specific entity in an image. {A capsulenet can discriminate the consistency of the contextual information of the capsules and therefore has a better ability to model the spatial positions of visual features of images than a CNN. Due to their advantages, capsulenets have been widely used in many image processing tasks, such as face recognition, optical character recognition, and scene classification \cite{MS-CapsNet_xiang,xiang2020matrix}. }

  In a capsulenet, the steps for transformation from the output active vector $u_i$ of the shallow feature layer to the output active vector $v_j$ of the high feature layer are as follows:\par
  First, the $u_i$ of the previous capsule is multiplied by a weighted matrix $W_{i j}$, and then the prediction vector $u_{j \mid i}$ can be obtained by
  \begin{equation}
   u_{j \mid i}=W_{i j} u_{i},
  \end{equation}
where $u_i$ represents the $i$th feature in the shallow layer, and $u_{j \mid i}$ is its $j$th prediction feature in the high feature layer. Then, $s_j$ in capsule $j$ is obtained as a weighted sum of all the prediction vectors $u_{j \mid i}$ from the shallow layer. 
\begin{equation}
s_{j}=\sum_{i} c_{i j} u_{j \mid i},
\end{equation}
where $c_{i j}$ denotes the coupling coefficient, which is determined by a routing softmax processing of dynamic routing \cite{Capsulenet_Sabour}, and $s_j$ is the input vector of capsule $j$. Finally, capsule $j$ uses the length of $s_j$ to determine the probability of the existence of the entity; therefore, a nonlinear function called squash function is used to squash vector $s_j$ as follows
\begin{equation}
v_{j}=\frac{\left\|s_{j}\right\|^{2}}{1+\left\|s_{j}\right\|^{2}} \frac{s_{j}}{\left\|s_{j}\right\|},
\end{equation}
where $v_{j}$ represents the output of capsule $j$, which can be considered a vector representation of the input. Through the capsulenet, a more robust extracted feature representation of the input image patch can be obtained. Later, the probability of the entity can be determined by calculating the length of the activity vector.

\section{The Proposed Approach }
\label{proposed approach}

Assume an input HSI $\mathbf X$ with the size of $H \times W \times D$, and $\mathbf X \in \mathbb{R}^{H \times W \times D}$. Each training pixel $\mathbf{X}^{k} \in \mathbb{R}^{1 \times1 \times D}$, $k=1,2,...,K$, where $D$ and $K$ represent the dimension of HSI and the number of training samples, respectively. Suppose that $T$ represents the total class in HSI. For each pixel  $\mathbf{X}^{k} $, the truth label of $\mathbf{X}^{k} $ can be represented as $\mathbf t^{k} $, which is actually a vector of length $t$ with value ``1'' at the position of the correct label and ``0'' elsewhere.  Different from the natural image classification, which inputs whole images, the HSI classification based on CNN uses image patches centered on labeled pixels as the input samples. {Here, we conduct a standard preprocessing step, i.e., the principal component analysis (PCA) algorithm,  for HSI dimension reduction.} Therefore, HSI's band channels shrink from $D$ to $d$. Let $\mathbf{X}^{k} \in \mathbb{R}^{1 \times1 \times d}$ be the center pixel of dimension reduction images, the image patch can be represented as $\mathbf{Y}^{k} \in \mathbb{R}^{m \times m \times d}$ of the window size $m \times m$, and $\{\mathbf{Y}^{k},\mathbf t^{k}\}^K_{k=1}$ represents the input samples. Since the ASP units are the core components of the proposed ASPCNet, in Section \ref{Adaptive Spatial Layer}, we first give a detailed description of the ASP unit and ASP-based convolution operation (ASPConvs) in the task of HSIC. Then, we develop the ASP unit into the process of the original capsulenet (ASPCaps) in Section \ref{Adaptive spatial unit for capsulenet}. Finally, the overall architecture of the proposed ASPCNet is given in Section \ref{Architecture of ASPCNet} and Fig. \ref{fig:framework}.
\begin{figure}
	\centering
	{\includegraphics[scale=0.67]{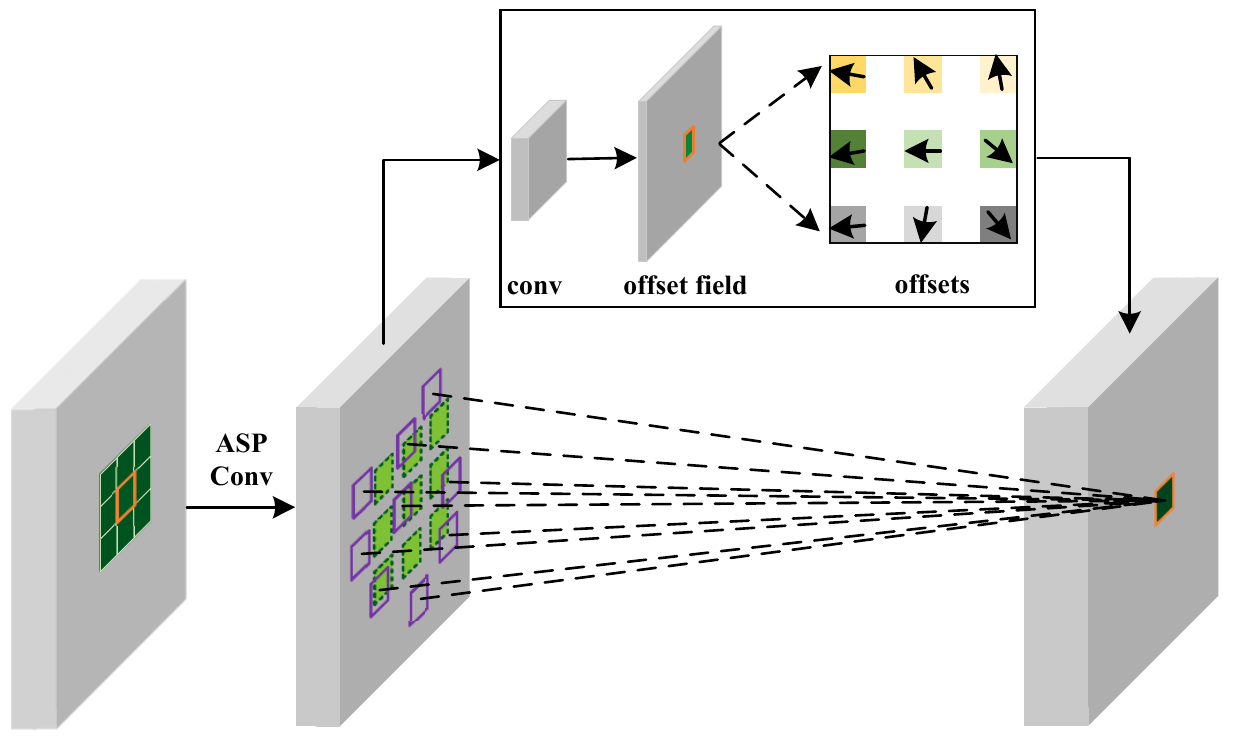}\par
	\caption{Illustration of the process of the ASPConv module. The pixel marked by the orange box is the central pixel.  After ASPConv is applied to the input images, the sampling locations of regular convolutional filters can be transferred to the pixels in the purple boxes.}
	\label{fig:adaptive_unit}}
\end{figure}

\subsection{ASPConvs}
\label{Adaptive Spatial Layer}

Given an image patch (centered on labeled pixel) $\mathbf{Y}^{k} \in \mathbb{R}^{m \times m \times d}$ with a window size of $m \times m$, the sampling locations of the input image can be represented as $s_1 = \{(0,0),(0,1),...,(m-1, m-1)\}$. The traditional convolution operation consists of two steps: 1) sampling using a series of regular grids $\Re$ of size $p \times q$ slide over the input image patch and 2) the summation of sampled values weighted by $\bf{W}$ for different convolution kernels. A larger size of regular grids $\Re$ can be replaced by multiple smaller sizes of $\Re$ with the same receptive field. As the convolution network model goes deeper, the pooling layers are supposed to shrink the model's parameters. Unfortunately, the principle behind the pooling operation is to preserve distinctive and representative features of each pooling area. This means that some details tend to disappear after several pooling operations, which is extremely unfavorable for classification accuracy. In addition, the sampling locations of the general convolution layer are fixed; this configuration cannot effectively extract the spatial information according to the different structures of input image patches. Therefore, to overcome these issues, the ASP unit is proposed in this paper, which simultaneously expands the receptive field and rotates the sampling location, i.e., integrating dilated and deformation operations into one unit (see Fig. \ref{fig:adaptive_unit}). The details are shown as follows:\par 

First, we can introduce the dilated convolution to implement the larger receptive field with fewer parameters. For instance, two 3$\times$3 filters along with a 2-dilation rate can be equally replaced by one 9$\times$9 filter with a 1-dilation rate, which can prevent the loss of the spatial relative relationship of pixels caused by the downsampling operation. Grid $\Re'$ defines the locations of the modified dilated convolution kernels, which are described below. Suppose the dilation rate $ h_{dr}$=3, $\Re'= \{(x,y) | (-3,-3),(-3,-1),...(1,3),(3,3)\}$. Here, $h_{dr}$ can be changed freely in this model according to the different shapes of images; its effect is analyzed in Section \ref{Analysis the influence of the dilation rate}. For any dilation rate $h_{dr}$, its size can be calculated as
\begin{equation}
p_{dr}=p+(p-1)(h_{dr}-1),
\end{equation}
\begin{equation}
q_{dr}=q+(q-1)(h_{dr}-1),
\end{equation}
where $p_{dr}$ and $q_{dr}$ refer to the height and width of a kernel of dilated convolution, respectively. There are only no more than $p$$\times$$q$ nonzero values in a $p_{dr}$$\times$$q_{dr}$ area.\par
\begin{figure}
	\centering
	{\includegraphics[scale=0.68]{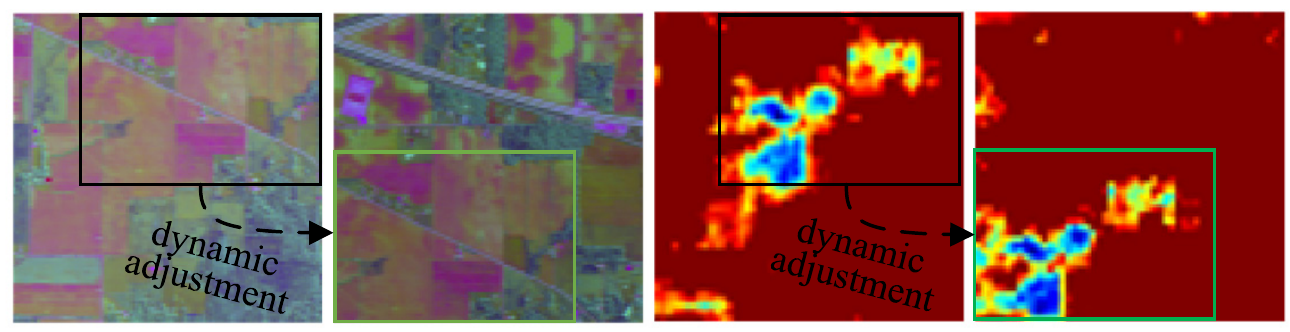}\par
	\hspace{0.4cm }{\footnotesize(a)}\hspace{1.8cm }{\footnotesize(b)}\hspace{ 2cm }{\footnotesize(c)}    \hspace{ 1.8cm }{\footnotesize(d)}
	\caption{Two image patches with the same region of interest. (a) and (b) Pseudo color images and (c) and (d) Visualization features of contributing scores. Different colors indicate the proportion of contribution in the feature learning process, where blue indicates higher contribution scores and red indicates lower contribution scores.}}
	\label{fig:attention}
\end{figure}
However, without a brilliantly designed dilated convolution structure \cite{wang2018understanding}, multiple cascaded dilated convolutional layers result in a griding effect while hurting the continuity of the kernel due to the checkerboard processing method. {Based on this observation, a deformable convolution operation is combined with dilated convolution to adaptively adjust the sampling location of the grid kernel.} Therefore, the new unit called the ASP unit is promoted, which can directly choose the grid location in an adaptive way to search for the most relevant pixels in a local area and achieve a larger receptive field, fewer parameters, and more spatial information. \par

Specifically, given one location ($x, y$) of the grid $\Re'$ and its value $\Re'(x,y)$ which corresponds to two values $\Delta x$, $\Delta y$ in the offset field. In this way, the adaptive spatial unit is generated to fuse the information of similar neighboring pixels, which can be obtained by 
\begin{equation}
\Re'_{\text {new }}(x, y)=\mathbf{Y}^{k}(\Re'(x, y)+(\Delta x, \Delta y)) = \Re'\left(x', y'\right),
\end{equation}
where $x'$ and $y'$ refer to fractional locations and can be represented as $x'=\min (\max (0, x+\Delta x), p-1)$, and {$y'=\min (\max (0, y+\Delta y), q-1)$}. The value of $\Re'(x', y')$ is calculated according to the values of the surrounding integer locations via bilinear interpolation. 

Finally, the adaptive spatial feature ${\bf u}(x, y)$ of location $(x, y)$ can be extracted as follows

\begin{equation}
{\bf u}(x, y)=\sum\limits_{i = 1}^{Len(\Re')} {w_{i} \cdot \Re'_{\text {new}} \left(x_{i}, y_{i}\right)\cdot\Delta m_i} ,
\end{equation}
where $w_{i}$ represents the corresponding weight of the kernel. $\Delta m_i$ refers to the weight of the $i$th sampling pixel, and if the sample pixel does not meet the sampling rules, $\Delta m_i$tends toward zero \cite{Deformable-v2}. $Len(\cdot)$ represents the length of the vector $\Re'$. We provide a visual explanation of the proposed ASPConv module via gradient guided back-propagation in Fig. \ref{fig:attention}. Focusing on the framed area, we can draw the conclusion that the proposed method can always adjust the shapes of the region of interest in adaptive rules according to the real objects. We also package the ASP unit into a simple and easy-to-use network layer that can replace the traditional convolutional layer and can be placed anywhere.

\subsection{ASPCaps}
\label{Adaptive spatial unit for capsulenet}

\begin{figure}
	\centering
	{\includegraphics[scale=0.615]{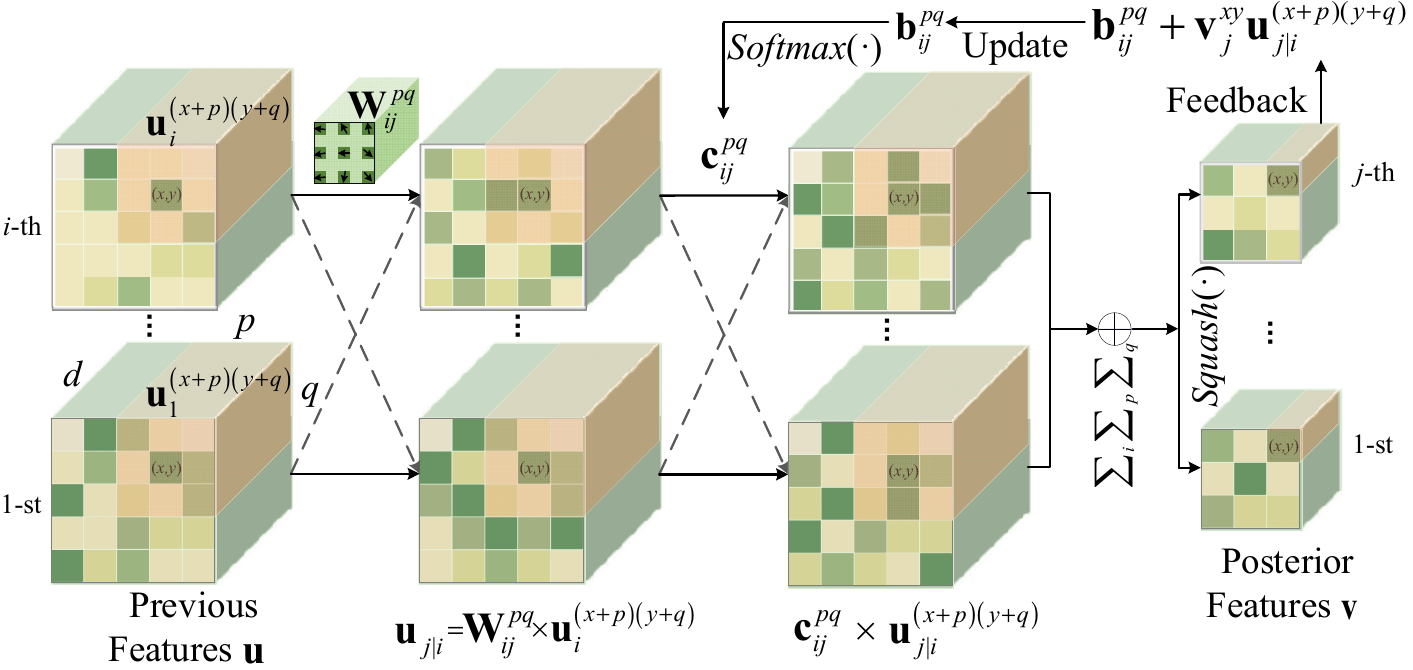}\par
	\caption{Illustration of the process of the ASPCaps module, which vividly describes the transfer process from previous features {\bf u} to posterior features {\bf v}.}
	\label{fig:capsule_stepsfigure}}
\end{figure}
To create a deep and robust capsulenet with fewer parameters, there are some considerations: 1)  Some robust capsule layers should be staked. 2) Location connections and shared routing matrices should be added to a dynamic routing algorithm. 3) More adaptive spatial information should be learned through the location connection. {Based on these considerations, we develop a novel deep capsulenet named ASPCNet with a special ASP unit.} Specifically, an ASP-based conv-capsule named ASPCaps is introduced during the dynamic routing process to replace the general conventional convolution layer. The detailed description is shown below.

After the ASPConvs module operation, the relatively high-level features ${\bf u}$ can be obtained as the input features to transfer into the primarycaps layer. The ASPCaps operation has the ability to explore the relationship among capsules. Fig. \ref{fig:capsule_stepsfigure} shows an illustration of the ASPCaps operation. Here, we divide the previous layer feature ${\bf u}$ into multiple capsules, where ${\bf{u}} = \left\{ {{{\bf{u}}_i},i = 1, \ldots ,I} \right\}$. Supporting the posterior layer feature $\bf v$ with $j$ capsules, the detailed transmission process is described in the following. For each capsule during the dynamic routing process, all the capsules in their receptive fields make a prediction through the transform matrix. Here, given a pixel at location $(x,y)$, ${\bf u}_{i}^{(x+p)(y+q)}$ represents the output of the capsule, which is the $i$-th capsule in the last capsule layer at position $(x+p, y+q)$, and ${\bf W}_{i j}^{p q}$ represents the shared transform matrix between the $i$th capsule of layer ${\bf u}$ and the $j$th capsule of layer $\bf v$, which possesses adaptive rotation rules based on ASPConv. $I$ is the number of capsules in the last capsule layer.

\begin{equation}
{\bf u}_{j \mid i}^{(x+p)(y+q)}={\bf W}_{i j}^{p q} {\bf u}_{i}^{(x+p)(y+q)},
\end{equation}
where ${\bf u}_{j \mid i}^{(x+p)(y+q)}$ denotes the $j$th prediction feature in the high feature layer. As all the ``prediction vectors'' ${\bf u}_{j \mid i}$ are obtained, the weighted sum of all capsules of $(x+p)(y+q)$ can serve as the input of the capsule. 

\begin{equation}
{\bf s}_{j}^{x y}=\sum_{i} \sum_{p} \sum_{q} {\bf c}_{i j}^{p q} {\bf u}_{j \mid i}^{(x+p)(y+q)},
\end{equation}
where ${\bf s}_{j}^{x y}$ is the input vector of capsule $j$ and ${\bf c}_{i j}^{p q}$ denotes the coupling coefficients, which are obtained by the softmax function and updated by the dynamic routing algorithm as follows
\begin{equation}
{\bf c}_{i j}^{pq}=\frac{\exp \left({\bf b}_{i j}^{pq}\right)}{\sum_{j=1}^{J} {\bf b}_{ij}^{pq}},
\end{equation}
where $J$ is the total number of capsules of posterior layer ${\bf v}$. ${\bf b}_{i j}^{pq}$ is initialized to 0 before the training begins and is determined by the dynamic routing algorithm. In the dynamic routing algorithm, the coefficient ${\bf b}_{i j}^{pq}$ is iteratively refined by measuring the agreement between the ``prediction vector'' ${\bf u}_{j \mid i}^{(x+p)(y+q)}$ and ${\bf v}_{j}^{x}$. If agreement is reached to a great extent, capsule ${\bf u}_{j \mid i}^{(x+p)(y+q)}$ make a good prediction for capsule ${\bf v}_{j}^{xy}$. Then, the coefficient ${\bf b}_{i j}^{pq}$ is significantly increased. In our network, the agreement is quantified as the inner product between two vectors ${\bf u}_{j \mid i}^{(x+p)(y+q)}$ and ${\bf v}_{j}^{xy}$. This agreement is added to ${\bf b}_{i j}^{pq}$ as follows 
\begin{equation}
  {\bf b}_{i j}^{pq} \leftarrow {\bf b}_{i j}^{pq}+ {\bf u}_{j \mid i}^{(x+p)(y+q)} {\bf v}_{j}^{xy}. 
\end{equation}

Finally, the input vector is squashed by a nonlinear function (i.e., the \emph{squash} function) to generate the output of the capsule. The detailed equations are listed below
\begin{equation}
{\bf v}_{j}^{x y}=\operatorname{Squash}\left({\bf s}_{j}^{x y}\right).
\end{equation}

\subsection{Architecture of the ASPCNet}% 参照dssnet
\label{Architecture of ASPCNet}

\begin{figure}
	\centering
	\includegraphics[scale=0.54]{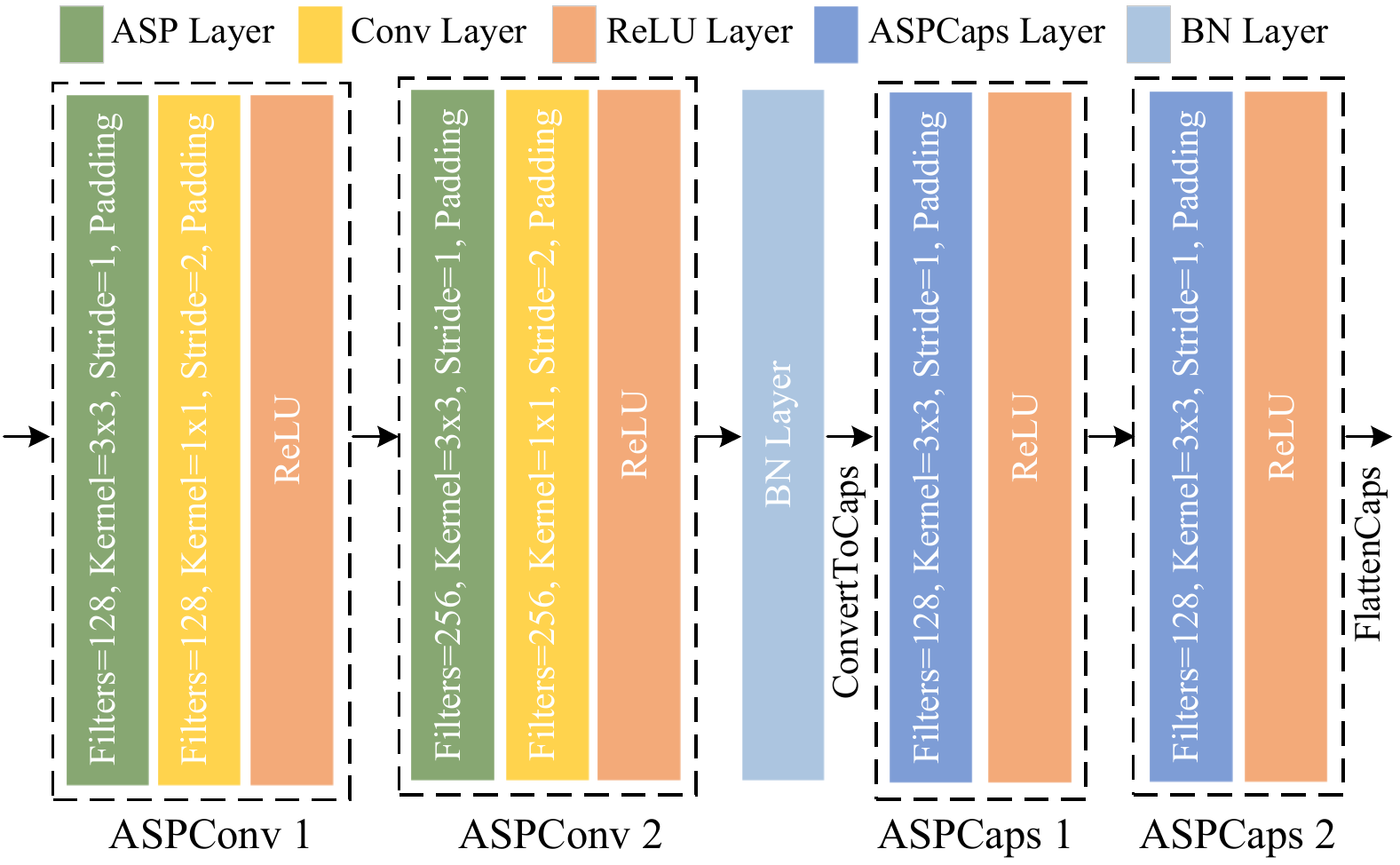}\par
	\caption{Network architecture of the proposed ASPCNet model.}
	\label{fig:smallframe}
\end{figure}
 
  In the following, we describe the whole framework of HSIC based on ASPCNet, which is shown in Fig. \ref{fig:framework}. The whole architecture can be divided into several steps: 1) Dimension reduction by using PCA. 2) Primary feature extraction by using ASPConvs modules. 3) Hierarchical structure extraction by using ASPCaps modules. 4) Image patch-based classification.  {Here, the proposed ASPCNet consists of four blocks: ASPConv 1, ASPConv 2, ASPCaps 1, and ASPCaps 2, which are shown in Fig. \ref{fig:smallframe}. For details on the configuration of these hyper parameters, we refer the reader to the APPENDIX section.}\par   

 After learning the deep ASPCNet features ${\bf v}$, we use a fully connected capsule layer to obtain digital capsule ${\bf v}'$, which is of size $T \times {16}$. Then, the Euclidean norm (i.e., $\left\| {{\ell _2}} \right\|$) is used to obtain the probability vector ${\bf v}''$ for each test patch with size $T \times 1$, where each value is mapped to [0, 1]. Once the deep network is well-trained, for a test patch $\mathbf{Y}^{test} \in \mathbb{R}^{m \times m \times d}$, the class label of its center pixel can be determined based on the maximum probability
\begin{equation}
\operatorname{Class}\left(\mathbf{Y}^{test}\right)=\underset{t=1,2, \ldots, T}{\arg \max } {\bf v}''_{t}.
\end{equation}

Here, we use the margin loss \cite{Capsulenet_Sabour} as the loss function in this paper, since it can increase the probability of true classes compared with that of the traditional cross-entropy loss. For each capsule $j$ in the last capsule layer, its loss function $ L_{j}$ can be calculated as follows
\begin{equation}
\begin{aligned}
L_{j}=&T_{j} \cdot \max \left(0, n^{+}-\left\|{\mathbf v}_{j}\right\|\right)^{2}\\
&+\lambda\left(1-T_{j}\right) \cdot \max \left(0,\left\|{\mathbf v}_{j}\right\|-n^{-}\right)^{2},
\end{aligned}
\end{equation}
where $T_{j}$ = $1$ when class $j$ is actually present and equals to $0$ otherwise. We set $n^{+}$ = 0.9 and $n^{-}$ = 0.1 as the lower bound and the upper bound for the correct class and the wrong class\cite{Capsulenet_Sabour} , respectively. $\lambda$ is a hyper-parameter that controls the effect of gradient backpropagation in the initial phase during training. The total loss of the model is the sum of the loss of all the output capsules of the last layer.

\section{Experiments results and analysis}
\label{experimental}
	In this section,  three experiments are performed on the Salinas, University of Pavia, and Houston datasets. The details of these three datasets are described as follows:

\subsection{Datasets}
\begin{figure}
	\centering
	\includegraphics[scale=0.6]{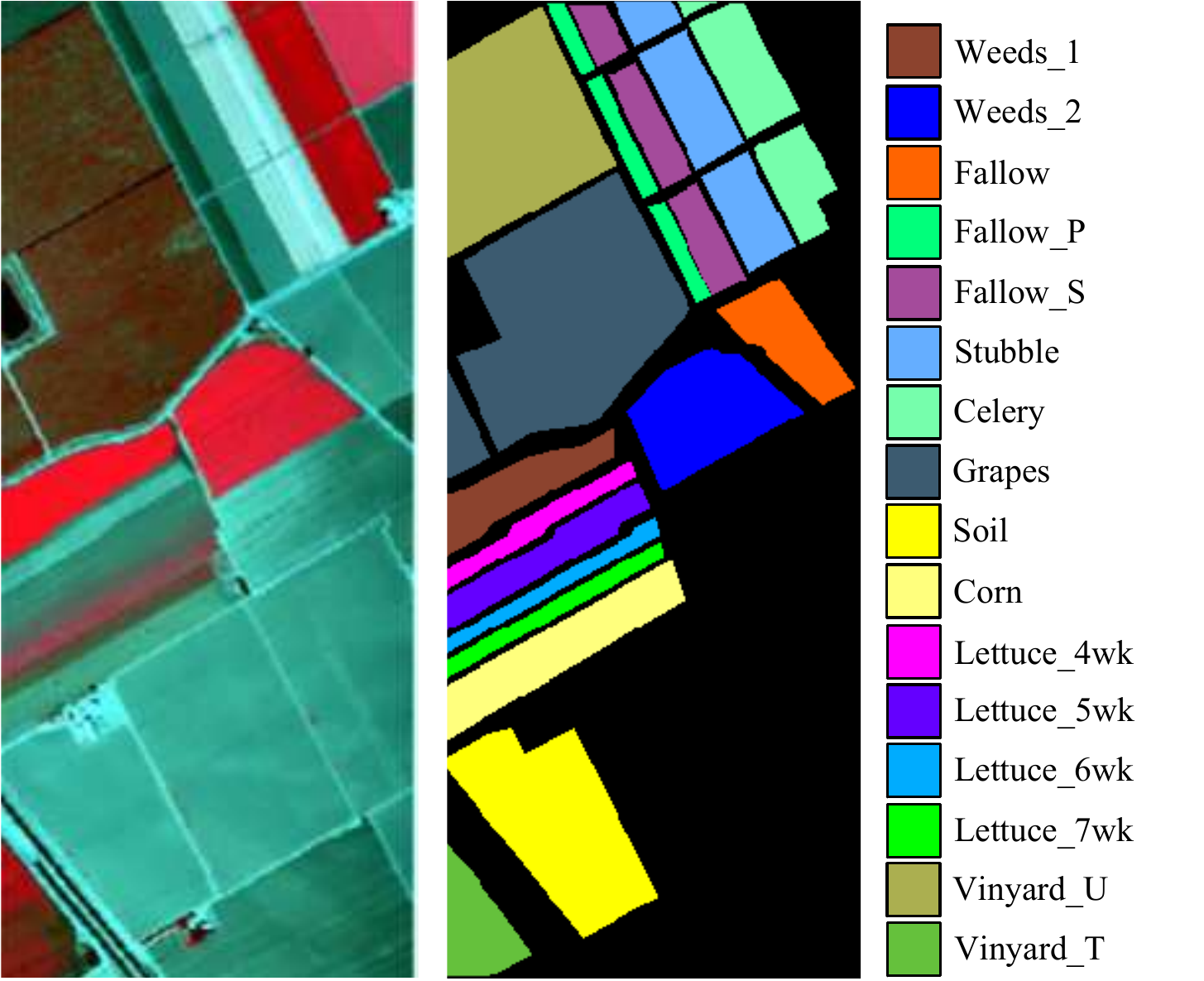}\par
	{\footnotesize(a)}\hspace{2.7cm }{\footnotesize(b)}\hspace{ 2.2cm }{\footnotesize(c)}    \hspace{ 0cm }
	\caption{The Salinas dataset. (a) Three-band color composite of the Salinas image. (b) and (c) the corresponding
	reference data.}
	\label{fig:salinas_gt}
\end{figure}

\begin{figure}
	\centering
	\includegraphics[scale=0.65]{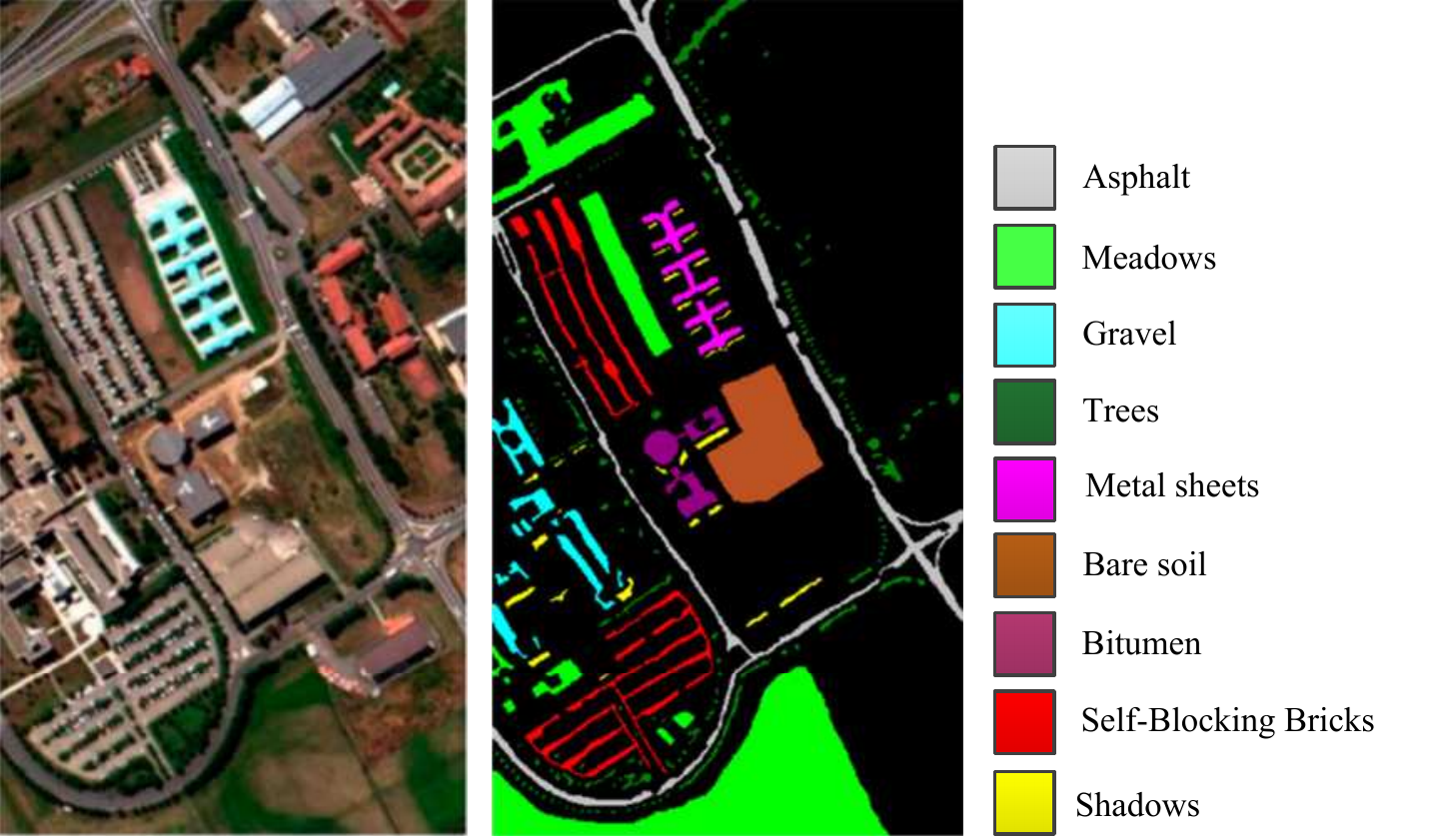}\par
	{\footnotesize(a)}\hspace{2.7cm }{\footnotesize(b)}\hspace{ 2.2cm }{\footnotesize(c)}    \hspace{ 0cm }
	\caption{The University of Pavia dataset. (a) Three-band color composite of the University of Pavia image. (b) and (c) the corresponding reference data.}
	\label{fig:PaviaU_gt}
\end{figure}

\begin{figure}
	\centering
    \subfigure[]{\includegraphics[scale=0.95]{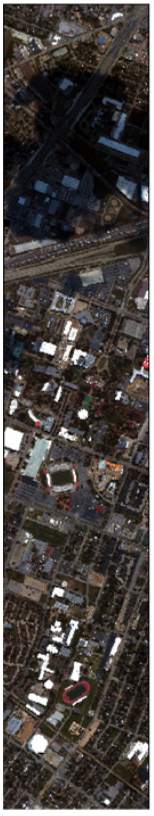}}\ \ \
    \subfigure[]{\includegraphics[scale=0.95]{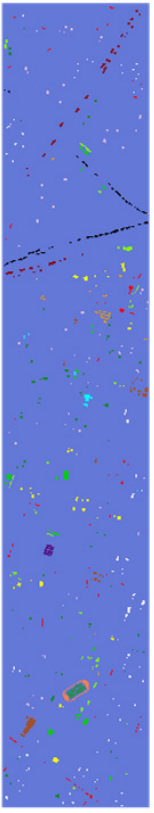}}\ \ \
    \subfigure[]{\includegraphics[scale=0.85]{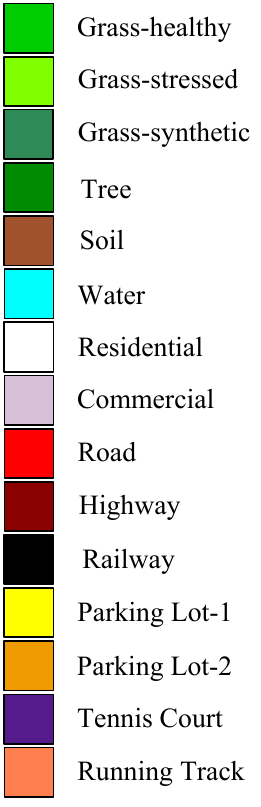}}\par
	\caption{The Houston dataset. (a) Three-band color composite of the Houston image. (b) and (c) the corresponding reference data.}
	\label{fig:hu_gt}
\end{figure}

\begin{enumerate}[{1)}]
	% \item {\em Indian Pines}: The first one namely Indian Pines image is an Airborne Visible/Infrared Imaging Spectrometer (AVIRIS )image, which shows the Indian Pines Test Field in the northwest of Indiana in 1992. The image is characterized by 145$\times$145 pixels, a spatial resolution of 20 m per pixel and 220 bands in the range of 0.2-2.4 $\mu$m.  With removing 20 water absorption wave bands, the remaining 200 bands are used in the experiment. Fig. \ref{fig:indian_gt} shows the  false-color composite of the Indian Pines image and the corresponding reference data with 16 different classes. \par

	\item {\em Salinas}: The first dataset is the Salinas image dataset, which was acquired by  AVIRIS over the Salinas Valley in southern California. {The images have 224 bands and 512$\times$217 pixels with a spatial resolution of 3.7 m. }According to the reference classification map shown in Fig. \ref{fig:salinas_gt}, there are 16 different classes labeled with different colors.\par

	\item {\em University of Pavia}: The second dataset is called the University of Pavia dataset and was photographed by a Reflective Optics System Imaging Spectrometer (ROSIS-3) in the range of 0.46-0.86 $\mu$m over the University of Pavia. The scenes contain 115 bands and 610$\times$340 pixels with a spatial resolution of 1.3 m per pixel. After 12 fairly noisy bands are removed, the experiments are performed using only 103 bands. Fig. \ref{fig:PaviaU_gt}(a-c) shows a false color composite of the University of Pavia image, the reference classification map, and the color labels, and there are 9 different classes.\par

	\item {\em  Houston}: The last image set, namely the Houston images, was captured by an ITRES-CASI 1500 sensor, and shows the University of Houston campus and its neighboring field. The images contain 349$\times$1905 pixels at a spatial resolution of 2.5 m per pixel and 144 bands in the range of 0.38-1.05 $\mu$m. Fig. \ref{fig:hu_gt} shows the  false-color composite of the Indian Pines image and the corresponding reference data with 15 different classes. \par
	\end{enumerate}\par

\subsection{Compared Methods}
\label{Compared Methods}
{In this subsection, we compare the proposed method to other methods, i.e., a SVM \cite{svm2004}, extreme learning machine (ELM) \cite{moreno2014extreme}, EPF \cite{EPF}, deep convolution neural network (DeepCNN), CapsuleNet \cite{paoletti2018capsule}, deformable convolution neural network (DHCNet) \cite{Deformable}, and spectral-spatial fully convolutional network (SSFCN) \cite{bigdata}.} The implementation details of each method included in the experiments are summarized as follows.

\begin{enumerate}[  1)]
\item SVM: Classification is performed with the original spectral features via an SVM classifier with an RBF kernel \cite{svm2004}. This algorithm is implemented using the LIBSVM library \cite{svm2004} and the parameters of the RBF-SVM are chosen using five-fold cross-validation.\par
\item ELM: Original spectral classification is performed with ELM \cite{moreno2014extreme}. This is a simple machine learning algorithm with only one hidden layer and one output layer and the default parameters of the ELM in \cite{moreno2014extreme} are adopted.\par
 
\item EPF: This is a spectral-spatial classification method with the first 20 principal components determined via guided filter-based EPF. We use the default parameters of a filtering size $r$ and blur degree $\epsilon$ to 3 and 0.2, respectively \cite{EPF}, in the implementations. \par

\item DeepCNN: Spatial classification is performed via a 3D-CNN method with the first 20 principal components. The architecture is shown in Table \ref{tab:DCNNArchitecture}. For each pixel, a patch size of 27$\times$27 is extracted as the input of the network. This algorithm is trained using the Adam optimizer with a learning rate of 5e-4.\par

%输入应该是

\begin{table}
\centering
  { \caption{The structure of the DeepCNN method. $d$ represents the image dimension, and $T$ is the total class of HSIs. The Stride of Convolution Layer and Maxpooling Layer Are Set to 1 And 2, Respectively.}
    \begin{tabular}{llll}
    \hline
    {Layer Name} & {Input Shape} & {Kernel Size} & {Output Shape} \\
    \hline
    \hline
    Input Layer & (27, 27, $d$)  & -     & (27, 27, $d$)  \\
    Conv Layer  & (27, 27, $d$)  & (3×3×$d$, 32)  & (27, 27, 32)    \\
    Conv Layer  & (27, 27, 32)  & (3×3×32, 32)  & (27, 27, 32)    \\
    Conv Layer  & (27, 27, 32)  & (3×3×32, 64) & (27, 27, 64)    \\
    Conv Layer  & (27, 27, 64)  & (3×3×64, 64) & (27, 27, 64)    \\
    Maxpooling & (27, 27, 64)    & (2×2)    & (14, 14, 64)    \\
    Conv Layer  & (14, 14, 64)  & (3×3×64, 128)  & (14, 14, 128)    \\
    Conv Layer & (14, 14, 128)  & (3×3×128, 128)   & (14, 14, 128)    \\
    Conv Layer & (14, 14, 128)  & (3×3×128, 256)   & (14, 14, 256)    \\
    Conv Layer & (14, 14, 256)  & (3×3×256, 256)   & (14, 14, 256)    \\
    Maxpooling & (14, 14, 256)    & (2×2)    & (7, 7, 256)    \\
    FC Layers    & (12544)    & -         & (1280)    \\
    FC Layers  & (1280)    & -      & (128)    \\
    FC Layers  & (128)    & -      & ($T$)        \\
    \hline
    \end{tabular}
  \label{tab:DCNNArchitecture}}
\end{table}

\begin{table}
  \centering
  {\caption{The structure of the proposed ASPCNet method. $d$ represents image dimension, and $T$ is the total class of HSI. The Stride of Convolution Layers are described in Appendices \ref{appendices}.}
    \begin{tabular}{llll}
    \hline
    {Layer Name} & {Input Shape} & {Kernel Size}  & {Output Shape} \\
    \hline
    \hline
    Input Layer & (27, 27, $d$)  & -       & (27, 27, $d$)  \\
    ASP Layer 1 & (27, 27, $d$)  & (3$\times$3$\times$$d$, 128)     & (27, 27, 128)    \\
    Conv Layer & (27, 27, 128)    & (1$\times$1$\times$128, 128)     & (14, 14, 128)  \\
    ASP Layer 2 & (14, 14, 128)  & (3$\times$3$\times$128, 256)    & (14, 14, 256) \\
    Conv Layer & (14, 14, 256)      & (1$\times$1$\times$256, 256)  & (7, 7, 256) \\
    BN Layer & (7, 7, 256) & -         & (7, 7, 256) \\
    ConvertToCaps & (7, 7, 256) & -       & (7, 7, 256, 1) \\
    ASPCaps Layer 1 & (7, 7, 256, 1) & (3$\times$3$\times$256, 32, 4)   & (7, 7, 32, 4)      \\
    ASPCaps Layer 2 & (7, 7, 32, 4) & (3$\times$3$\times$128, 32, 4)   & (7, 7, 32, 4) \\
    FlattenCaps & (7, 7, 32, 4) & -        & (1568, 4)    \\
    DigitalCaps & (1568, 4)    & ($T$, 1568, 16, 4)    & ($T$, 16)    \\
    CapsToScalars & ($T$, 16)    & -      & ($T$, 1)        \\
    \hline
    \end{tabular}%
  \label{tab:thenetworkofourmethod}}%
\end{table}%
\item CapsuleNet: The original capsule network-based HSIC method with the original spectral information \cite{paoletti2018capsule}. The architecture adopts the Adam optimizer with a learning rate equal to 0.001, 100 training epochs, and a patch size of 15$\times$15.\par

\item DHCNet: This is a deformable convolution neural network for HSIC with the first 3 principal components. According to the implementations \cite{Deformable}, the patch size, the momentum of the batch normalization layer, and the training epoch are set to 29$\times$29, 0.9, and 1500, respectively.\par

\item SSFCN: This is a spectral-spatial FCN with conditional random fields. The detailed architecture and default parameters of the network follow the implementations in \cite{bigdata}. The network is trained with a learning rate of 5e-4 using the Adam optimizer.\par

\item ASPCNet: This is the proposed spectral-spatial ASPC neural network. The detailed architecture of the network is shown in Table \ref{tab:thenetworkofourmethod}. {We set the batch size and training epoch to 96 and 200, respectively.} The momentum of the batch normalization layer is set to 0.9. The network is trained using the Adam optimizer with a learning rate of 5e-4, beta\_1 of 0.9, beta\_2 of 0.999, and epsilon of 1e-8.\par
\end{enumerate}

All of the methods are carried out on a desktop computer with a Windows 10 OS, an Intel i9-10900F 2.8-GHz processor with 64 GB of RAM, and a single NVIDIA GTX2080Ti GPU. DeepCNN, CapsuleNet, DHCNet, SSFCN, and ASPCNet are constructed by using the Keras framework, CUDA 10, and Python 3.6.5 as the programming language. The SVM, ELM, and EPF methods are carried out in MATLAB R2020a without GPU acceleration.

\begin{table}
  \centering
  \caption{Numbers of Training and Testing Samples in Salinas dataset.}
    \begin{tabular}{c|ccc}
        \hline
    Class Name & Total & Training & Testing \\
    \hline
    \hline
    \cellcolor[rgb]{ .549,  .263,  .18} {\textcolor[rgb]{ 1,  1, 1}{Brocoli green weeds 1}} & 2009 & 40 & 1969 \\
    \cellcolor[rgb]{ 0,  0,  1} {\textcolor[rgb]{ 1,  1, 1}{Brocoli green weeds 2}} & 3726 & 76 & 3650 \\
    \cellcolor[rgb]{ 1,  .392,  0} {Fallow } & 1976 & 38 & 1938 \\
    \cellcolor[rgb]{ 0,  1,  .522} {Fallow rough plow} & 1394 & 26 & 1368 \\
    \cellcolor[rgb]{ .643,  .294,  .608} {\textcolor[rgb]{ 1,  1,  1}{Fallow smooth}} & 2678 & 52 & 2626 \\
    \cellcolor[rgb]{ .396,  .682,  1} {Stubble} & 3959 & 79 & 3880 \\
    \cellcolor[rgb]{ .463,  .996,  .675} {Celery} & 3579 & 70 & 3509 \\
    \cellcolor[rgb]{ .235,  .357,  .439} {\textcolor[rgb]{ 1,  1,  1}{Grapes untrained}} & 11271 & 225 & 11046 \\
    \cellcolor[rgb]{ 1,  1,  0} {Soil vinyard develop} & 6203 & 124 & 6079 \\
    \cellcolor[rgb]{ 1,  1,  .49} {Corn senesced weeds} & 3278 & 21 & 3257 \\
    \cellcolor[rgb]{ 1,  0,  1} {Lettuce romaine 4wk} & 1068 & 21 & 1047 \\
    \cellcolor[rgb]{ .392,  0,  1} {\textcolor[rgb]{ 1,  1,  1}{Lettuce romaine 5wk}} & 1927 & 38 & 1889 \\
    \cellcolor[rgb]{ 0,  .675,  .996} {Lettuce romaine 6wk} & 916 & 18 & 898 \\
    \cellcolor[rgb]{ 0,  1,  0} {Lettuce romaine 7wk} & 1070 & 20 & 1050 \\
    \cellcolor[rgb]{ .671,  .686,  .314} {Vinyard untrained} & 7268 & 140 & 7128 \\
    \cellcolor[rgb]{ .396,  .757,  .235} {Vinyard treils} & 1807 & 36 & 1771 \\
    \hline
    \hline
    Total & 54129 & 1024  & 53105 \\
    \hline
    \end{tabular}%
  \label{tab:SAnumber}%
\end{table}%

\begin{table}[htbp]
  \centering
  \caption{Numbers of Training and Testing Samples in University of Pavia dataset.}
    \begin{tabular}{c|ccc}
    \hline
    Class Name & Total & Training & Testing \\
    \hline
    \hline
    \cellcolor[rgb]{ .753,  .753,  .753} {Asphalt} & 6631 & 200 & 6431 \\
    \cellcolor[rgb]{ 0,  1,  0} {Meadows} & 18649 & 200 & 18449 \\
    \cellcolor[rgb]{ 0,  1,  1} {Gravel} & 2099 & 200 & 1899 \\
    \cellcolor[rgb]{ 0,  .502,  0} {\textcolor[rgb]{ 1,  1,  1}{Trees}} & 3064 & 200 & 2864 \\
    \cellcolor[rgb]{ .996,  0,  1} {Metal sheets} & 1345 & 200 & 1145 \\
    \cellcolor[rgb]{ .647,  .322,  .161} {\textcolor[rgb]{ 1,  1,  1}{Bare soil}} & 5029 & 200 & 4829 \\
    \cellcolor[rgb]{ .502,  0,  .502} {\textcolor[rgb]{ 1,  1,  1}{Bitumen}} & 1330 & 200 & 1130 \\
    \cellcolor[rgb]{ 1,  0,  0} {Self-Blocking Bricks} & 3682 & 200 & 3482 \\
    \cellcolor[rgb]{ 1,  1,  0} {Shadows} & 947 & 200 & 747 \\
    \hline
    \hline
    Total & 42776 & 1800  & 40976 \\
    \hline
    \end{tabular}%
  \label{tab:PUnumber}%
\end{table}%

\begin{table}[htbp]
  \centering
  \caption{Numbers of Training and Testing Samples in Houston dataset.}
    \begin{tabular}{c|ccc}
    \hline
    Class Name & Total & Training & Testing \\
    \hline
    \hline
    \cellcolor[rgb]{ 0,  .804,  0} Grass-healthy & 1251 & 160 & 1091 \\
    \cellcolor[rgb]{ .498,  .98,  1} Grass-stressed & 1254 & 160 & 1094 \\
    \cellcolor[rgb]{ .18,  .545,  .341} \textcolor[rgb]{ 1,  1,  1}{Grass-synthetic} & 697 & 80 & 617 \\
    \cellcolor[rgb]{ 0,  .545,  0} \textcolor[rgb]{ 1,  1,  1}{Tree}  & 1244 & 160 & 1084 \\
    \cellcolor[rgb]{ .627,  .322,  .176} \textcolor[rgb]{ 1,  1,  1}{Soil}  & 1242 & 160 & 1082 \\
    \cellcolor[rgb]{ 0,  1,  1} Water & 325 & 80 & 245 \\
    Residential & 1268  & 160   & 1108 \\
    \cellcolor[rgb]{ .847,  .749,  .847} Commercial & 1244 & 160 & 1084 \\
    \cellcolor[rgb]{ 1,  0,  0} Road  & 1252 & 160 & 1092 \\
    \cellcolor[rgb]{ .545,  0,  0} \textcolor[rgb]{ 1,  1,  1}{Highway} & 1227 & 160 & 1067 \\
    \cellcolor[rgb]{ 0,  0,  0} \textcolor[rgb]{ 1,  1,  1}{Railway} & 1235 & 160 & 1075 \\
    \cellcolor[rgb]{ 1,  1,  0} Parking Lot-1 & 1233 & 160 & 1073 \\
    \cellcolor[rgb]{ .933,  .604,  0} Parking Lot-2 & 469 & 80 & 389 \\
    \cellcolor[rgb]{ .333,  .102,  .545} \textcolor[rgb]{ 1,  1,  1}{Tennis Court} & 428 & 80 & 348 \\
    \cellcolor[rgb]{ 1,  .498,  .314} Running Track & 660 & 80 & 580 \\
    \hline
    \hline
    Total & 15029 & 2000  & 13029 \\
    \hline
    \end{tabular}%
  \label{tab:HUnumber}%
\end{table}%

\subsection{Quantitative Metrics}
To effectively evaluate the classification performance of different methods, three objective indexes, i.e., the overall accuracy (OA), average accuracy (AA), and Kappa coefficient (Kappa) are adopted in the experiments.  Specifically, the OA value represents the percentage of all test pixels that are correctly classified. the AA value measures the mean of all class accuracies.  Taking the uncertainty factors of the classification into consideration, the Kappa value is proposed, which represents the percentage of correctly classified pixels corrected by the degree of agreement.

\subsection{Analysis of the influences of different parameters}
\label{Analysis the influences of different parameters}
In this subsection, a detailed analysis of the influence of three parameters (i.e., the number of dimensions $d$, patch size $m$$\times$$m$ and dilation rate $h_{dr}$) on the image classification performance is conducted on three real datasets.

\begin{figure*}
	\centering{
	\subfigure[]{\includegraphics[scale=0.67]{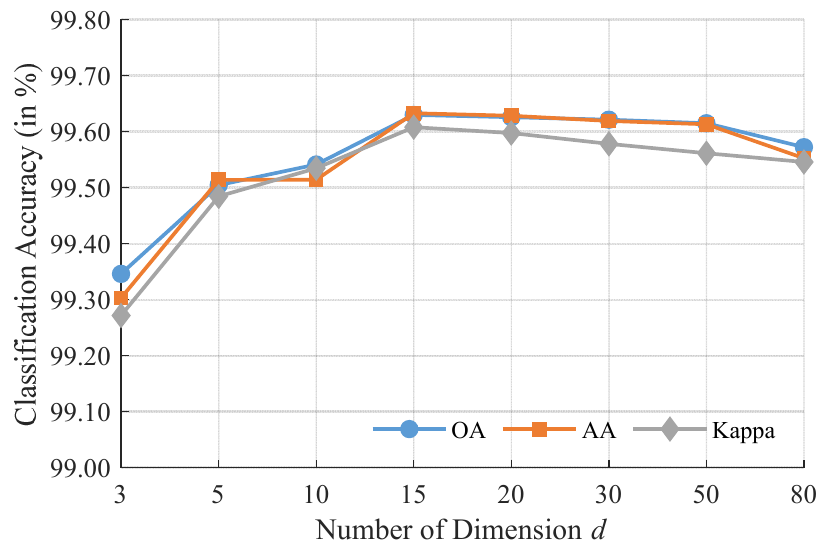}}
	\subfigure[]{\includegraphics[scale=0.67]{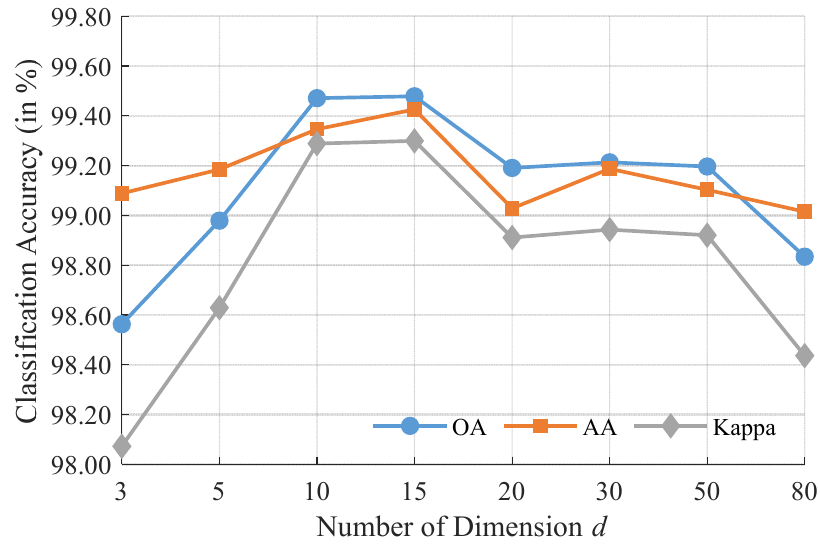}}
	\subfigure[]{\includegraphics[scale=0.67]{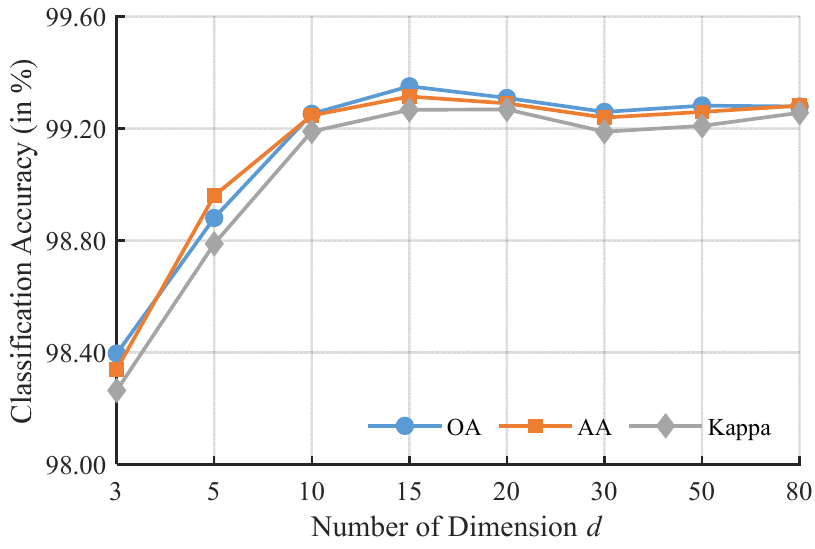}}\par
	\caption{The effects of the number of dimension $d$ based on the proposed method for three real datasets: (a) Salinas, (b) University of Pavia, (c) Houston.}
	\label{fig:Dim}}
\end{figure*}

\begin{figure*}
	\centering{
	\subfigure[]{\includegraphics[scale=0.67]{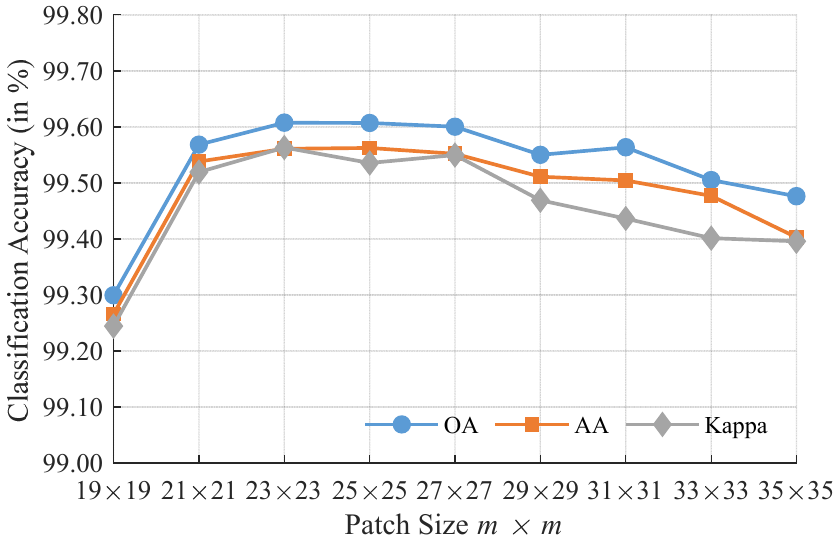}}
	\subfigure[]{\includegraphics[scale=0.67]{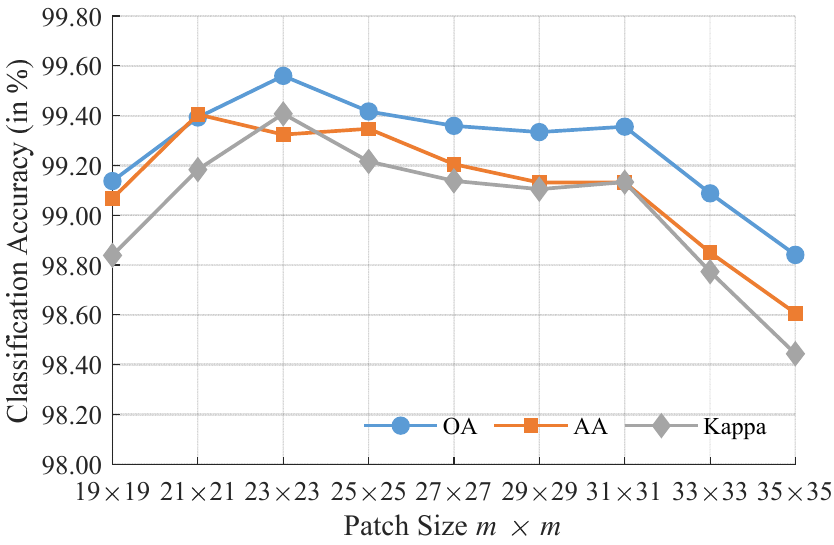}}
	\subfigure[]{\includegraphics[scale=0.67]{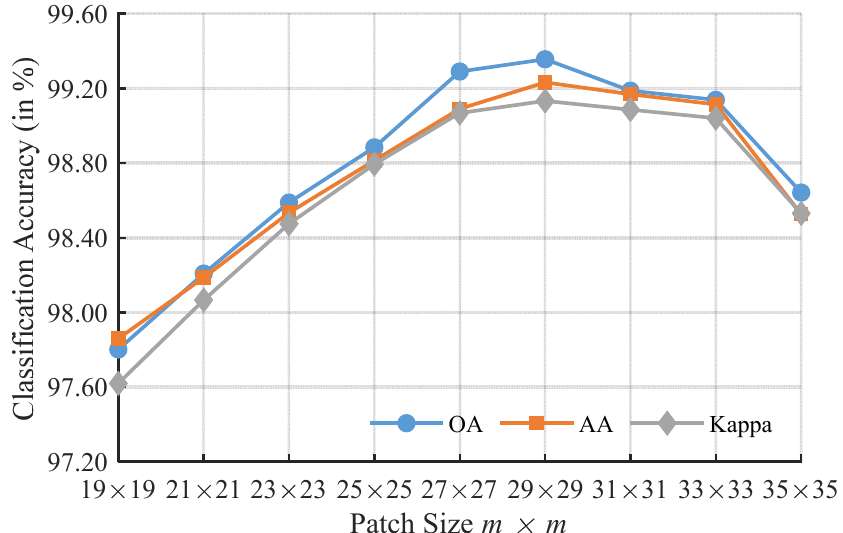}}\par
	\caption{The effects of the patch size $m$$\times$$m$ based on the proposed method for three real datasets: (a) Salinas, (b) University of Pavia, (c) Houston.}
	\label{fig:patchsize}}
\end{figure*}

\begin{figure*}
	\centering{
	\subfigure[]{\includegraphics[scale=0.67]{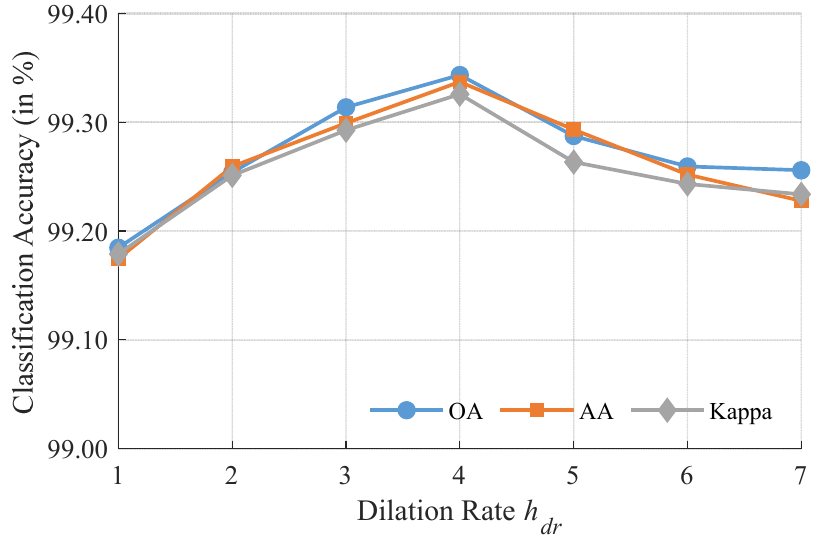}}
	\subfigure[]{\includegraphics[scale=0.67]{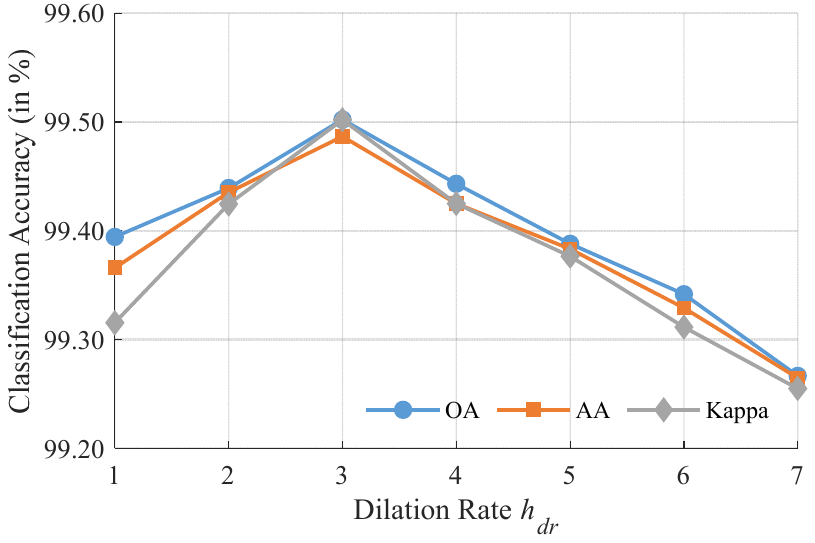}}
	\subfigure[]{\includegraphics[scale=0.67]{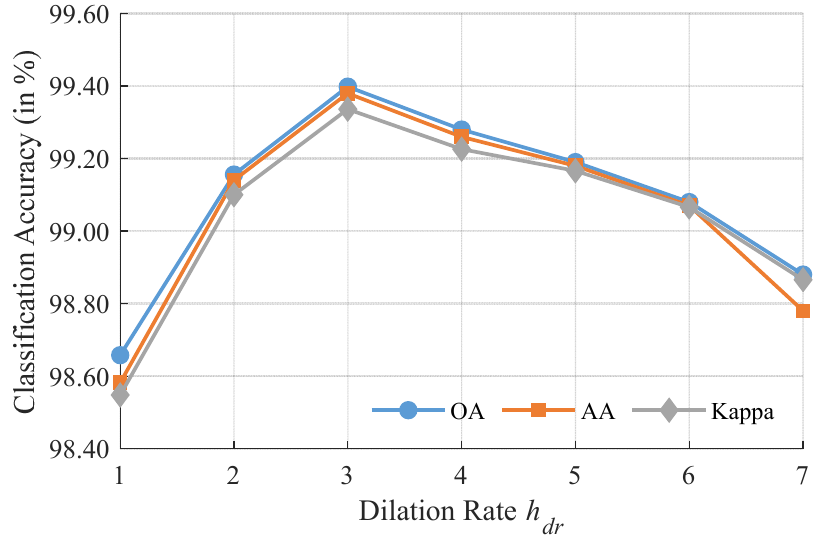}}\par
	\caption{The effects of the dilation rate $h_{dr}$ based on the proposed method for three real datasets: (a) Salinas, (b) University of Pavia, (c) Houston.}
	\label{fig:dilationrate}}
\end{figure*}
\subsubsection{Analysis of the influence of the number of dimensions}	
first, we analyze the influence of the number of dimensions (the principal components) of HSI under different settings.  Fig. \ref{fig:Dim} shows the influence of parameter $d$ on three different datasets, i.e.,  the Salinas, University of Pavia, and Houston datasets. As seen in Fig. \ref{fig:Dim}(a), for the Salinas dataset, the classification performance of the ASPCNet method increases first and then decreases slightly when $d >$ 15 as parameter $d$ increases because the first several components include most of the spatial information. With an increasing number of components, there is much extra even noisy information among the HSI bands, and this situation is even worse for a complex dataset such as the University of Pavia (see Fig. \ref{fig:Dim}(b)). In addition, a similar phenomenon can be seen on the Houston datasets in Fig. \ref{fig:Dim}(c). Therefore, $d$ = 15 is set as the default parameter among all datasets.

\begin{figure*}
	\centering{
	\subfigure[]{\includegraphics[scale=0.67]{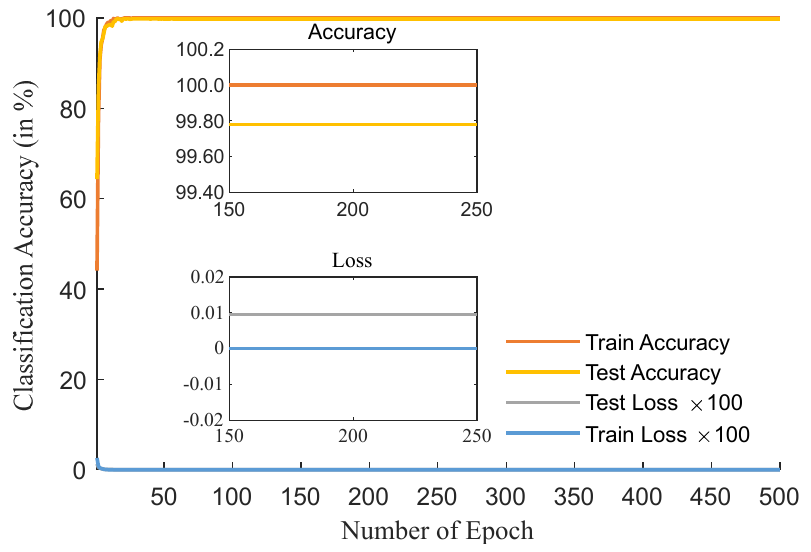}}
	\subfigure[]{\includegraphics[scale=0.67]{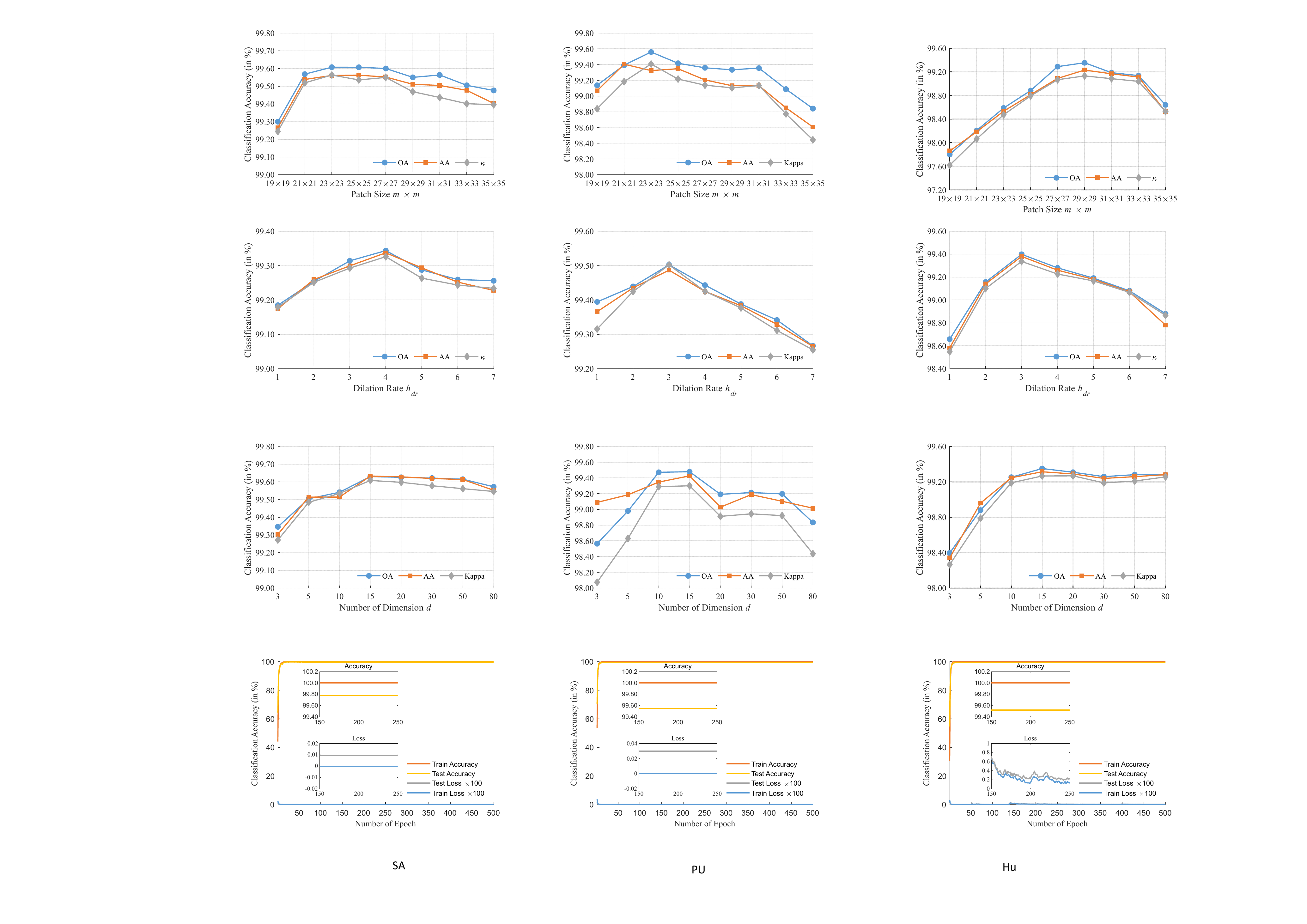}}
	\subfigure[]{\includegraphics[scale=0.67]{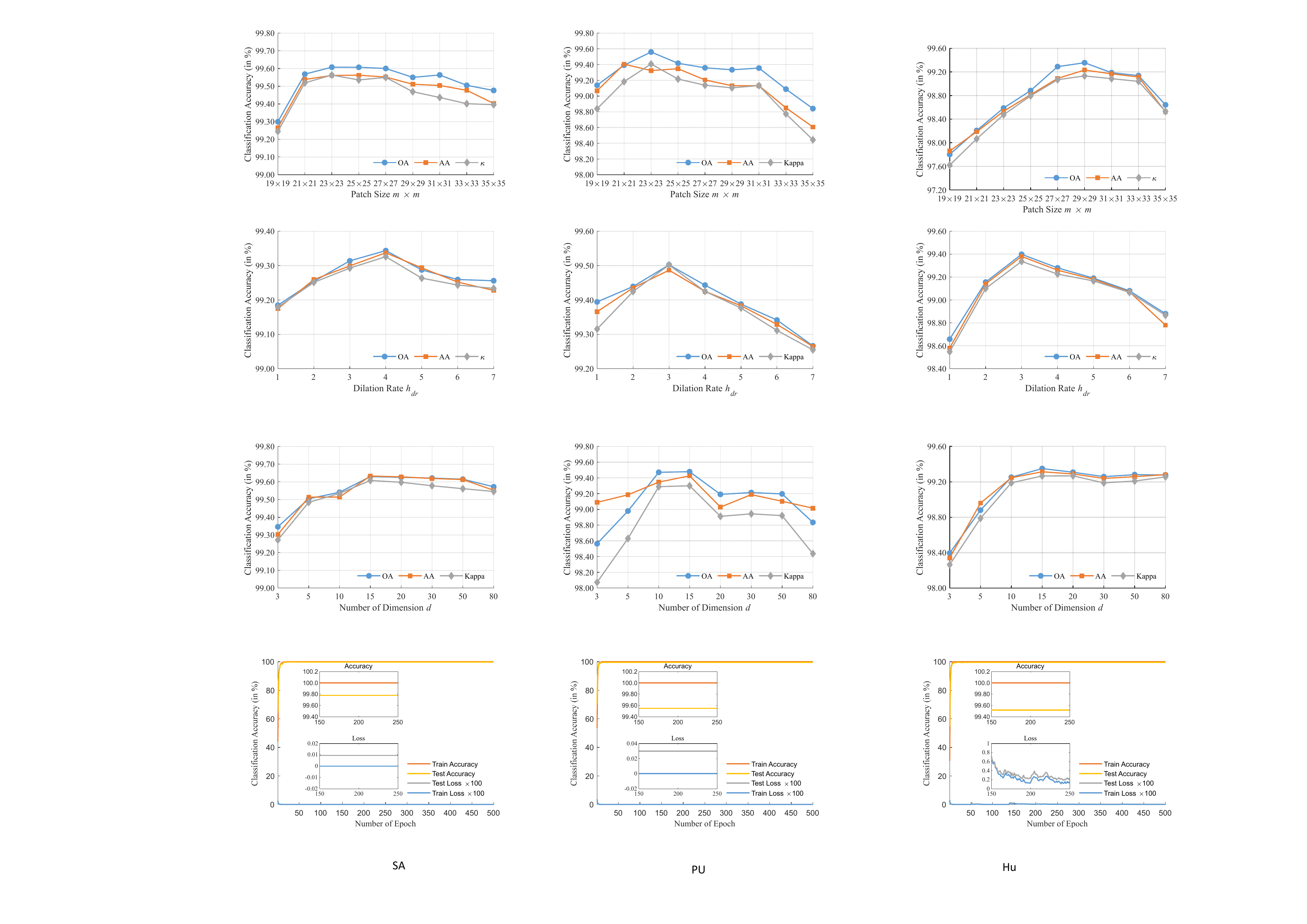}}\par
	% \subfigure[]{\includegraphics[scale=0.8]{./Figures/Epoch/IN_Epoch}}	
	\caption{The effects of the training epoch based on the proposed method for three real datasets. (a) Salinas, (b) University of Pavia, (c) Houston.}
	\label{fig:Epoch}}
\end{figure*}
\subsubsection{Analysis of the influence of the patch size}
second, we discuss the influence of the patch size $m$ of OA under different settings. Experiments are conducted on the  three datasets. For instance, Fig. \ref{fig:patchsize}(a) denotes different patch sizes on the Salinas dataset, showing the OA values versus different settings of patch size $m$ varying from 19$\times$19 to 35$\times$35 with an interval of 2. As seen, as the parameter $m$ increases, the OA increases first and declines later. Experimental results indicate that it can  extract more local spatial features at first. As $m$ increases (e.g., 35$\times$35), it may introduce extra information as interference. The same conclusion is reached for the other two datasets. Therefore, in consideration of the trade-off between classification performance and the number of parameters, a patch size of 27$\times$27 is adopted as the default parameter.

\begin{table*}
  \centering
   {
  \caption{Classification Accuracies (\%) on Salinas dataset Among Different Methods. Number in Parenthesis Indicates the Standard Variance of the Repeated Experiments. The Best and Second Results Are Shown in Red and Blue colors, Respectively.}
    \begin{tabular}{cc||cccccccc}
    \hline
    \multicolumn{2}{c||}{Classes} & SVM   & ELM   & EPF   &  DeepCNN  & CapsuleNet & DHCNet & SSFCN & ASPCNet \\
    \hline
    \hline
    \multicolumn{1}{c|}{\multirow{16}[2]{*}{CA}}& \cellcolor[rgb]{ .549,  .263,  .18} \textcolor[rgb]{ 1,  1,  1}{1} & 99.94(0.08) & 99.95(0.10) & \textcolor[rgb]{ 1,  0,  0}{\textbf{100.00(0.0)}} &  99.08(0.95) & 99.90(1.35) & 95.99(2.89) & 99.81(0.39) & \textcolor[rgb]{ 0,  0,  1}{\textbf{99.96(0.06)}} \\
    \multicolumn{1}{c|}{} & \cellcolor[rgb]{ 0,  0,  1} \textcolor[rgb]{ 1,  1,  1}{2} & 99.07(0.41) & 99.24(0.34) & 99.94(0.12) & 99.69(0.61) & 99.95(0.46) & 98.76(0.89) & \textcolor[rgb]{ 1,  0,  0}{\textbf{100.00(0.0)}} & \textcolor[rgb]{ 0,  0,  1}{\textbf{99.99(0.01)}} \\
    \multicolumn{1}{c|}{} & \cellcolor[rgb]{ 1,  .392,  0} 3     & 93.54(1.02) & 97.16(0.33) & 95.99(0.77) & 99.87(0.12) & 99.48(0.27) & 99.55(0.41) & \textcolor[rgb]{ 1,  0,  0}{\textbf{100.00(0.0)}} & \textcolor[rgb]{ 1,  0,  0}{\textbf{100.00(0.0)}} \\
    \multicolumn{1}{c|}{} & \cellcolor[rgb]{ 0,  1,  .522} 4     & 97.36(1.19) & \textcolor[rgb]{ 0,  0,  1}{\textbf{99.13(0.53)}} & 98.17(0.38) & 95.54(4.69) & 98.10(0.60) & 98.51(1.59) & \textcolor[rgb]{ 1,  0,  0}{\textbf{99.53(0.51)}} & 98.51(1.11) \\
    \multicolumn{1}{c|}{} & \cellcolor[rgb]{ .643,  .294,  .608} \textcolor[rgb]{ 1,  1,  1}{5} & 98.29(0.71) & 98.58(0.48) & \textcolor[rgb]{ 1,  0,  0}{\textbf{99.91(0.06)}} & 98.83(1.34) & \textcolor[rgb]{ 0,  0,  1}{\textbf{99.89(1.04)}} & 99.86(0.17) & 99.22(0.55) & 99.08(0.93) \\
    \multicolumn{1}{c|}{} & \cellcolor[rgb]{ .396,  .682,  1} 6     & 99.91(0.13) & \textcolor[rgb]{ 0,  0,  1}{\textbf{99.98(0.01)}} & \textcolor[rgb]{ 0,  0,  1}{\textbf{99.98(0.02)}} &  99.12(1.14) & 99.82(0.27) & 99.59(0.75) & \textcolor[rgb]{ 1,  0,  0}{\textbf{100.00(0.0)}} & \textcolor[rgb]{ 0,  0,  1}{\textbf{99.98(0.18)}} \\
    \multicolumn{1}{c|}{} & \cellcolor[rgb]{ .463,  .996,  .675} 7     & 99.16(0.75) & 99.73(0.13) & \textcolor[rgb]{ 0,  0,  1}{\textbf{99.91(0.12)}} & 99.77(0.36) & 99.00(0.64) & 99.56(0.52) & 99.86(0.07) & \textcolor[rgb]{ 1,  0,  0}{\textbf{100.00(0.0)}} \\
    \multicolumn{1}{c|}{}&\cellcolor[rgb]{ .235,  .357,  .439} \textcolor[rgb]{ 1,  1,  1}{8} & 76.53(1.39) & 79.72(0.77) & 80.67(2.04) & 99.27(0.55) & \textcolor[rgb]{ 0,  0,  1}{\textbf{99.77(1.07)}} & 98.86(0.93) & 98.23(0.82) & \textcolor[rgb]{ 1,  0,  0}{\textbf{99.89(0.19)}} \\
    \multicolumn{1}{c|}{} & \cellcolor[rgb]{ 1,  1,  0} 9     & 99.02(0.23) & 98.70(0.24) & 99.33(0.03) & \textcolor[rgb]{ 1,  0,  0}{\textbf{100.00(0.0)}} & 99.11(0.33) & \textcolor[rgb]{ 0,  0,  1}{\textbf{99.94(0.11)}} & \textcolor[rgb]{ 1,  0,  0}{\textbf{100.00(0.0)}} & \textcolor[rgb]{ 1,  0,  0}{\textbf{100.00(0.0)}} \\
    \multicolumn{1}{c|}{} & \cellcolor[rgb]{ 1,  1,  .49} 10    & 94.42(1.56) & \textcolor[rgb]{ 0,  0,  1}{\textbf{98.56(0.52)}} & 98.20(1.03) &  83.32(10.8) & 87.50(2.30) & 97.45(3.63) & 97.65(1.66) & \textcolor[rgb]{ 1,  0,  0}{\textbf{99.06(0.34)}} \\
    \multicolumn{1}{c|}{} & \cellcolor[rgb]{ 1,  0,  1} 11    & 90.66(3.72) & 93.96(1.35) & 96.73(1.45) & 94.96(3.25) & 98.57(3.48) & 99.58(0.36) & \textcolor[rgb]{ 1,  0,  0}{\textbf{99.92(0.15)}} & \textcolor[rgb]{ 0,  0,  1}{\textbf{99.68(0.32)}} \\
    \multicolumn{1}{c|}{} & \cellcolor[rgb]{ .392,  0,  1} \textcolor[rgb]{ 1,  1,  1}{12} & 95.30(0.78) & 94.94(0.75) & 99.09(0.33) & 97.49(2.68) & 99.41(0.65) & 99.93(0.09) & \textcolor[rgb]{ 1,  0,  0}{\textbf{99.98(0.03)}} & \textcolor[rgb]{ 0,  0,  1}{\textbf{99.96(0.18)}} \\
    \multicolumn{1}{c|}{} & \cellcolor[rgb]{ 0,  .675,  .996} 13    & 95.91(1.27) & 94.33(3.13) & 98.73(1.37) & 97.57(2.84) & 97.66(1.08) & \textcolor[rgb]{ 0,  0,  1}{\textbf{99.60(0.55)}} & \textcolor[rgb]{ 1,  0,  0}{\textbf{99.91(0.08)}} & 97.43(2.97) \\
    \multicolumn{1}{c|}{} & \cellcolor[rgb]{ 0,  1,  0} 14    & 95.33(0.80) & 96.94(1.63) & 98.63(0.88) & 97.77(1.03) & 98.38(1.66) & \textcolor[rgb]{ 1,  0,  0}{\textbf{99.73(0.30)}} & 99.20(0.53) & \textcolor[rgb]{ 0,  0,  1}{\textbf{99.43(0.82)}} \\
    \multicolumn{1}{c|}{} & \cellcolor[rgb]{ .671,  .686,  .314} 15    & 74.65(2.73) & 79.07(1.30) & 89.73(2.26) &  \textcolor[rgb]{ 1,  0,  0}{\textbf{99.78(0.21)}} & 97.95(0.97) & \textcolor[rgb]{ 0,  0,  1}{\textbf{99.56(0.30)}} & 98.38(0.47) & 99.10(1.08) \\
    \multicolumn{1}{c|}{} & \cellcolor[rgb]{ .396,  .757,  .235} 16    & 98.31(0.38) & 99.61(0.18) & \textcolor[rgb]{ 0,  0,  1}{\textbf{99.98(0.04)}} & 99.77(0.32) & 97.63(1.35) & 98.05(1.31) & 99.63(0.20) & \textcolor[rgb]{ 1,  0,  0}{\textbf{100.00(0.0)}} \\
    \hline
    \hline
    \multicolumn{2}{c||}{OA}    & 90.04(0.67) & 91.96(0.15) & 93.71(0.69) &  98.20(0.57) & 98.48(0.47) & 99.10(0.17) & 99.17(0.14) & \textcolor[rgb]{ 1,  0,  0}{\textbf{99.68(0.16)}} \\
    \multicolumn{2}{c||}{AA}    & 94.21(0.51) & 95.60(0.19) & 97.25(0.27) &  97.61(0.58) & 98.30(0.36) & 99.03(0.22) & \textcolor[rgb]{ 0,  0,  1}{\textbf{99.36(0.09)}} & \textcolor[rgb]{ 1,  0,  0}{\textbf{99.50(0.21)}} \\
    \multicolumn{2}{c||}{Kappa}     & 88.89(0.75) & 91.02(0.17) & 92.98(0.77) & 98.00(0.63) & 98.31(0.52) & 99.00(0.19) & 99.08(0.15) & \textcolor[rgb]{ 1,  0,  0}{\textbf{99.53(0.18)}} \\
    \hline
    \end{tabular}
  \label{tab:SAresult}}%
\end{table*}%
\subsubsection{Analysis of the influence of the dilation rate}
\label{Analysis the influence of the dilation rate}
the third experiment is conducted on the three datasets to evaluate the classification performance under different dilation rate settings. In general, a CNN uses a pooling operation to increase the receptive field. Unfortunately, some materials may "disappear" after several pooling layers. Dilated convolutions can avoid this problem in an elegant way, in which the dilation rate $h_{dr}$ exerts control. Here, Fig. \ref{fig:dilationrate} shows the influence of the dilation rate $h_{dr}$ on the three different datasets. As we can see from Fig. \ref{fig:dilationrate}(a), as the dilation rate $h_{dr}$ increases, OA increases first when $h_{dr} <$ 4 and then decreases when $h_{dr} >$ 4 on the Salinas dataset. For the University of Pavia and Houston datasets,  shown in Fig. \ref{fig:dilationrate}(b) and (c),  the proposed ASPCNet can obtain the highest accuracies when $h_{dr}$ equals 3. Therefore,  for the Salinas, University of Pavia and Houston datasets, the default parameters are set as $h_{dr}$ = 4, $h_{dr}$ = 3, and $h_{dr}$ = 3, respectively. 

Finally, we analyze the impact of the training epoch on the accuracies and losses of training and testing. From Fig. \ref{fig:Epoch}, it can be found that the proposed ASPCNet method can quickly converge with little oscillation. For example, it can always reach the maximum training accuracies when the epoch $\in$ [25, 30]. Therefore, if we want to improve the training efficiency in practical engineering applications, we can use the ``{\em early-stopping}'' operation; that is, when the training accuracies reach the local maximum and there are no better training accuracies in the subsequent training epochs, the training processing is automatically stopped.

\begin{figure*}
	\centering{
    \subfigure[]{\includegraphics[scale=0.6]{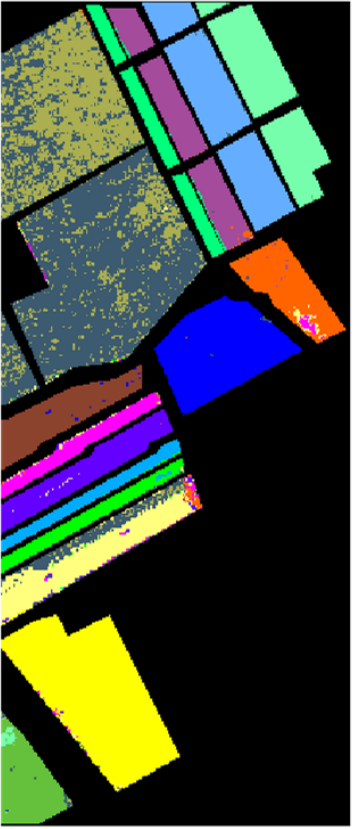}}
    \subfigure[]{\includegraphics[scale=0.6]{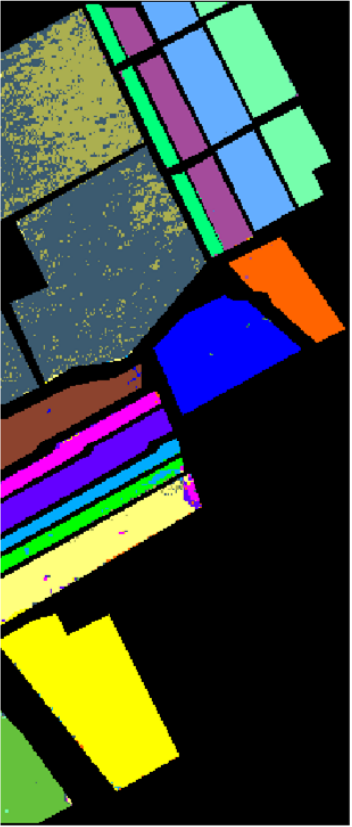}}
    \subfigure[]{\includegraphics[scale=0.6]{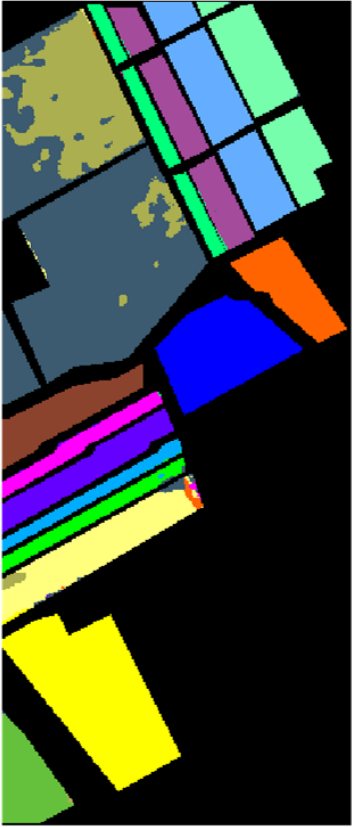}}
    \subfigure[]{\includegraphics[scale=0.6]{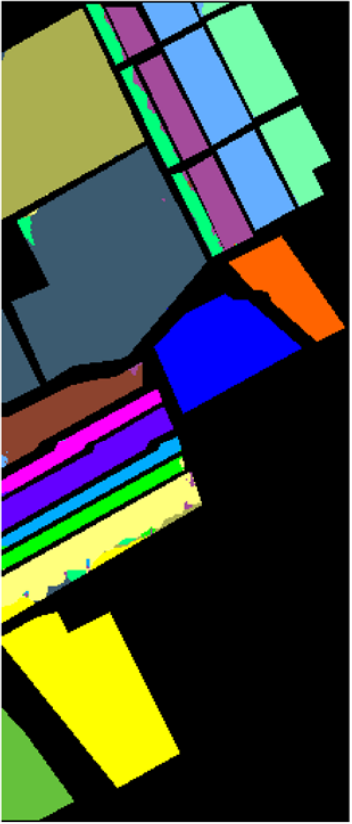}}
    \subfigure[]{\includegraphics[scale=0.6]{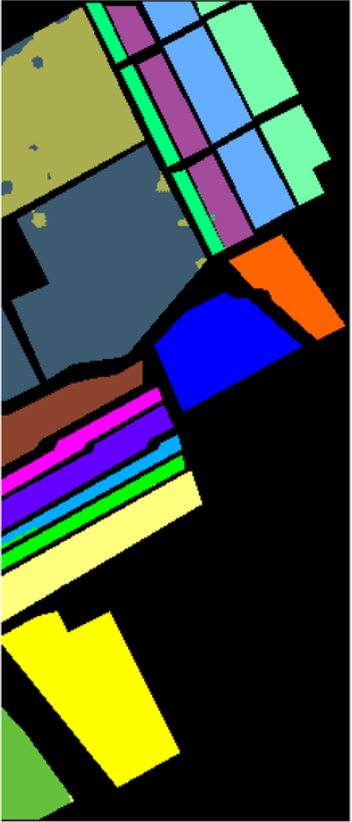}}
    \subfigure[]{\includegraphics[scale=0.6]{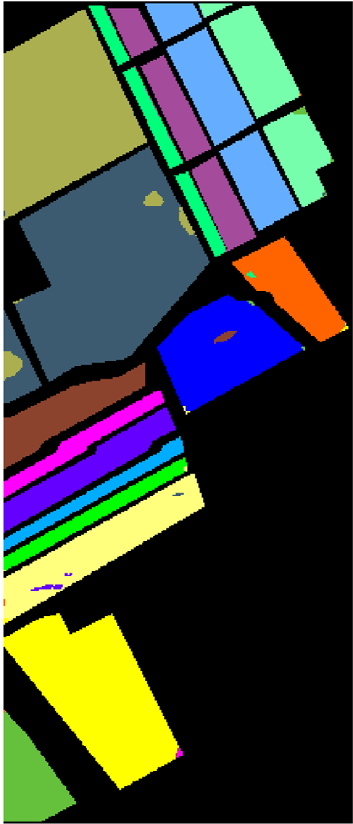}}
    \subfigure[]{\includegraphics[scale=0.6]{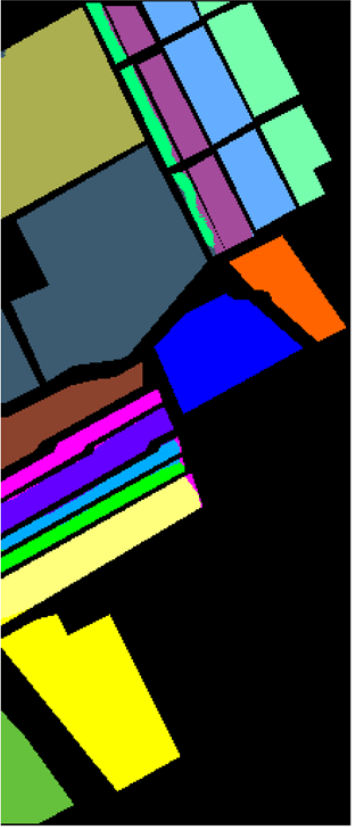}}
    \subfigure[]{\includegraphics[scale=0.6]{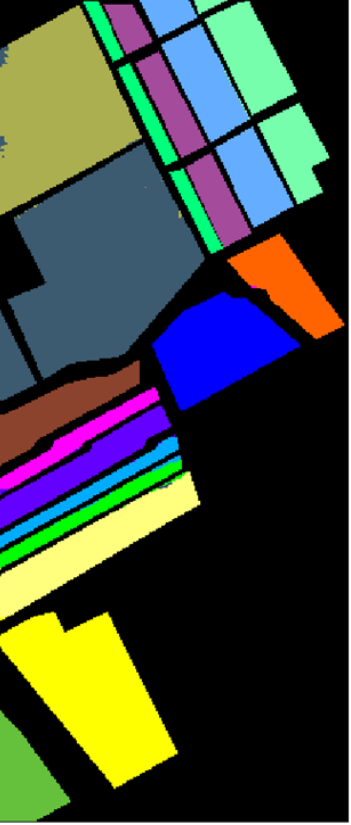}}\par
	\caption{Classification maps obtained by nine methods on the Salinas dataset: {(a)} SVM (90.04\%),  {(b)} ELM (91.96\%), {(c)} EPF (93.71\%), {(d)} DeepCNN (98.20\%), {(e)} CapsuleNet (98.48\%), {(f)} DHCNet (99.10\%), {(g)} SSFCN (99.17\%), and {(h)} ASPCNet (99.68\%). }
	\label{fig:SAresult}}
\end{figure*}

\begin{table*}
  \centering
   {
  \caption{Classification Accuracies (\%) on University of Pavia dataset Among Different Methods. Number in Parenthesis Indicates the Standard Variance of the Repeated Experiments. The Best and Second Results Are Shown in Red and Blue colors, Respectively.}
  \begin{tabular}{cc||cccccccc}
    \hline
       \multicolumn{2}{c||}{Classes}   & SVM   & ELM   & EPF    & DeepCNN  & CapsuleNet & DHCNet & SSFCN & ASPCNet \\
    \hline
    \hline
    \multicolumn{1}{c|}{\multirow{9}[2]{*}{CA}} &\cellcolor[rgb]{ .753,  .753,  .753} 1     & 97.15(0.22) & 97.46(0.44) & 98.46(0.32) &  97.16(2.00) & 97.20(1.21) & \textcolor[rgb]{ 0,  0,  1}{\textbf{98.66(0.73)}} & 98.54(0.99) & \textcolor[rgb]{ 1,  0,  0}{\textbf{99.97(0.04)}} \\
    \multicolumn{1}{c|}{} &\cellcolor[rgb]{ 0,  1,  0} 2     & 97.67(0.29) & 96.96(0.28) & 99.52(0.15) & 98.87(0.47) & 98.91(0.77) & 99.09(0.25) & \textcolor[rgb]{ 1,  0,  0}{\textbf{99.75(0.32)}} & \textcolor[rgb]{ 1,  0,  0}{\textbf{99.75(0.26)}} \\
    \multicolumn{1}{c|}{} &\cellcolor[rgb]{ 0,  1,  1} 3     & 77.89(1.35) & 69.34(1.55) & 96.16(2.07) & 99.25(1.31) & 98.05(0.83) & \textcolor[rgb]{ 0,  0,  1}{\textbf{99.49(0.69)}} & 98.85(1.28) & \textcolor[rgb]{ 1,  0,  0}{\textbf{99.84(0.18)}} \\
    \multicolumn{1}{c|}{} &\cellcolor[rgb]{ 0,  .502,  0} \textcolor[rgb]{ 1,  1,  1}{4} & 87.88(3.28) & 91.14(0.54) & 98.79(1.46) & 92.86(3.75) & \textcolor[rgb]{ 0,  0,  1}{\textbf{98.87(0.43)}} & \textcolor[rgb]{ 1,  0,  0}{\textbf{99.28(0.29)}} & 97.81(0.70) & 98.14(0.63) \\
    \multicolumn{1}{c|}{} &\cellcolor[rgb]{ .996,  0,  1} 5     & 97.61(0.97) & 99.41(0.19) & 99.53(0.60) & 99.55(0.49) & 99.74(0.20) & \textcolor[rgb]{ 0,  0,  1}{\textbf{99.88(0.17)}} & 99.23(0.20) & \textcolor[rgb]{ 1,  0,  0}{\textbf{100.00(0.0)}} \\
   \multicolumn{1}{c|}{} & \cellcolor[rgb]{ .647,  .322,  .161} \textcolor[rgb]{ 1,  1,  1}{6} & 79.82(3.44) & 77.44(1.31) & 95.59(2.04) & 99.79(0.40) & 99.66(0.26) & \textcolor[rgb]{ 0,  0,  1}{\textbf{99.97(0.03)}} & 99.69(0.04) & \textcolor[rgb]{ 1,  0,  0}{\textbf{100.00(0.0)}} \\
    \multicolumn{1}{c|}{} &\cellcolor[rgb]{ .502,  0,  .502} \textcolor[rgb]{ 1,  1,  1}{7} & 66.69(3.64) & 63.11(2.04) & 92.97(8.57) &  \textcolor[rgb]{ 0,  0,  1}{\textbf{99.98(0.04)}} & 98.55(0.57) & \textcolor[rgb]{ 0,  0,  1}{\textbf{99.98(0.04)}} & 99.77(0.08) & \textcolor[rgb]{ 1,  0,  0}{\textbf{100.00(0.0)}} \\
    \multicolumn{1}{c|}{} &\cellcolor[rgb]{ 1,  0,  0} 8     & 86.76(0.87) & 81.31(1.17) & 93.60(0.64) &  97.85(0.82) & 98.64(0.40) & \textcolor[rgb]{ 0,  0,  1}{\textbf{99.59(0.24)}} & 98.03(0.72) & \textcolor[rgb]{ 1,  0,  0}{\textbf{99.73(0.28)}} \\
    \multicolumn{1}{c|}{} &\cellcolor[rgb]{ 1,  1,  0} 9     & 99.84(0.10) & \textcolor[rgb]{ 0,  0,  1}{\textbf{99.97(0.05)}} & 98.95(0.95) & 98.15(0.91) & \textcolor[rgb]{ 1,  0,  0}{\textbf{100.00(0.0)}} & 98.90(0.49) & 99.23(0.72) & 98.76(0.52) \\
    \hline
    \hline
    \multicolumn{2}{c||}{OA}   & 88.24(0.72) & 89.70(0.37) & 97.91(0.39) & 98.26(0.46) & 98.70(0.56) & 99.25(0.16) & \textcolor[rgb]{ 0,  0,  1}{\textbf{99.28(0.24)}} & \textcolor[rgb]{ 1,  0,  0}{\textbf{99.58(0.13)}} \\
    \multicolumn{2}{c||}{AA}    & 87.92(0.79) & 86.24(0.37) & 97.06(0.95) & 98.16(0.61) & 98.85(0.35) & \textcolor[rgb]{ 0,  0,  1}{\textbf{99.23(0.10)}} & 99.14(0.25) & \textcolor[rgb]{ 1,  0,  0}{\textbf{99.55(0.09)}} \\
    \multicolumn{2}{c||}{Kappa}     & 88.37(0.91) & 86.34(0.48) & 97.19(0.53) & 97.66(0.61) & 98.25(0.75) & 98.99(0.21) & \textcolor[rgb]{ 0,  0,  1}{\textbf{99.23(0.33)}} & \textcolor[rgb]{ 1,  0,  0}{\textbf{99.44(0.13)}} \\
    \hline
    \end{tabular}%
  \label{tab:PUresult}}%
\end{table*}

\begin{figure*}
	\centering{
    \subfigure[]{\includegraphics[scale=0.50]{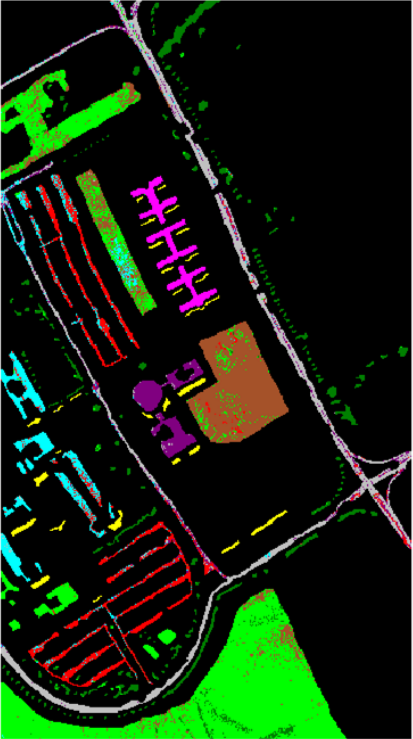}}
    \subfigure[]{\includegraphics[scale=0.50]{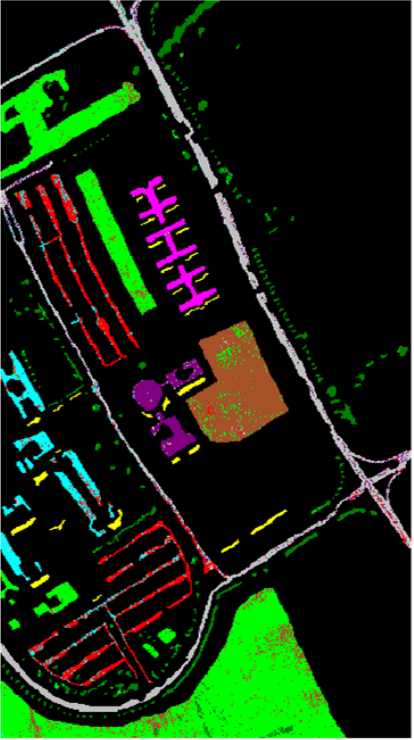}}
    \subfigure[]{\includegraphics[scale=0.50]{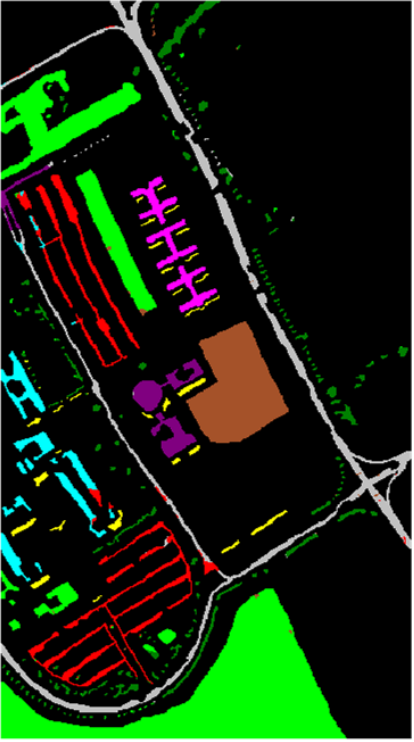}}
    \subfigure[]{\includegraphics[scale=0.50]{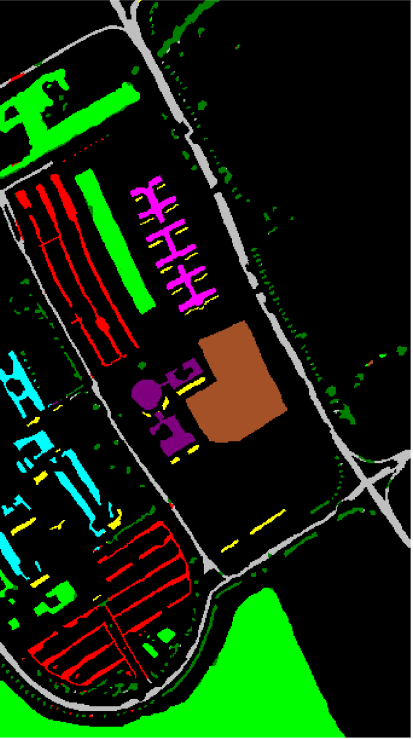}}
    \subfigure[]{\includegraphics[scale=0.50]{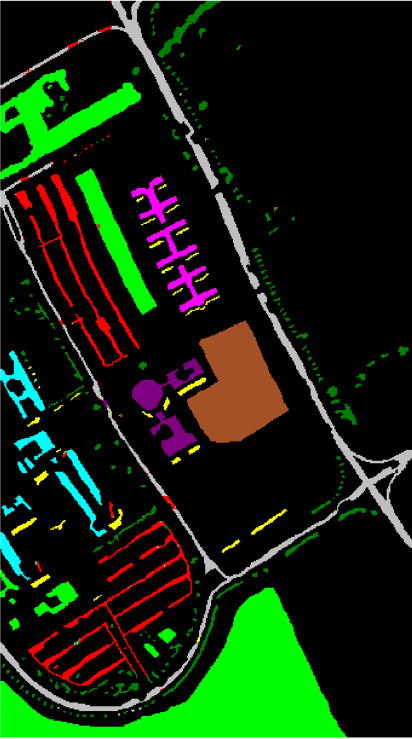}}
    \subfigure[]{\includegraphics[scale=0.50]{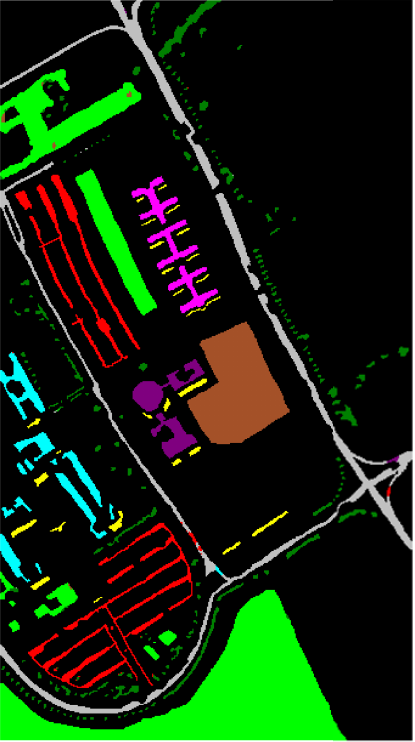}}
    \subfigure[]{\includegraphics[scale=0.50]{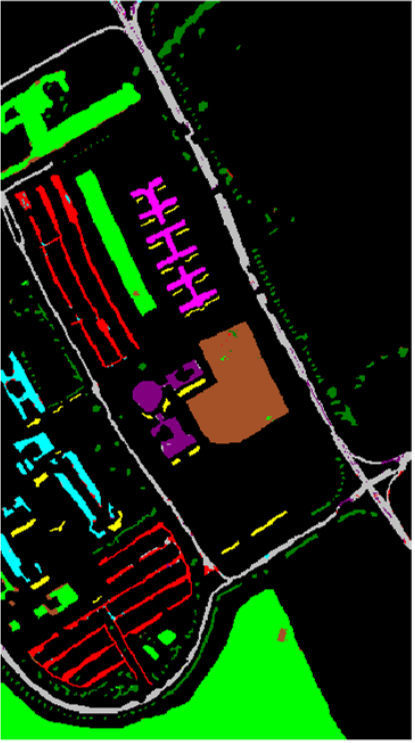}}
    \subfigure[]{\includegraphics[scale=0.50]{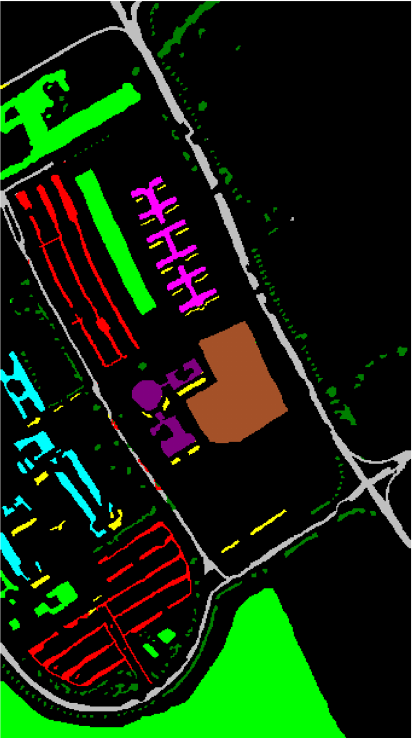}}\par
	\caption{Classification maps obtained by nine methods on the University of Pavia dataset: {(a)} SVM (88.24\%),  {(b)} ELM (89.70\%), {(c)} EPF (97.91\%), {(d)} DeepCNN (98.26\%), {(e)} CapsuleNet (98.70\%), {(f)} DHCNet (99.25\%), {(g)} SSFCN (99.28\%), and {(h)} ASPCNet (99.58\%).}
	\label{fig:PUresult}}
\end{figure*}
\subsection{Classification Results}
In this section, the proposed method named ASPCNet is compared with different well-known classification methods, including traditional machine learning-based methods such as SVM \cite{svm2004}, ELM \cite{moreno2014extreme}, and EPF \cite{EPF} and state-of-the-art (SOTA) deep learning-based methods such as DeepCNN, CapsuleNet \cite{paoletti2018capsule}, DHCNet \cite{Deformable}, and SSFCN \cite{bigdata}, on the Salinas, University of Pavia and Houston datasets. To evaluate the aforementioned three quantitative metrics, i.e., OA, AA and the Kappa, each experiment is repeated approximately 20 times for each classification method.\par

The first experiment is conducted on the Salinas dataset. Here,  2\% of the labeled image patches are randomly chosen as the training patches to train the parameters of the network and the rest of the patches are viewed as test patches, as seen in Table \ref{tab:SAnumber}.  The classification performance and the corresponding classification maps obtained by the nine methods are exhibited in Table \ref{tab:SAresult} and Fig. \ref{fig:SAresult}. The numbers in parentheses indicate the standard variances, and the  best and second best results are highlighted in red and blue, respectively. The SVM classifier obtains the worst classification accuracy on the Salinas dataset, which indicates that only using spectral information is not enough for complex HSIs. By taking the spatial information of HSI into consideration, traditional machine learning-based methods such as EPF can achieve better results with 93.71\%. Furthermore, image patch-based classifiers such as DeepCNN and CapsuleNet, which both obtain greater than 98\% classification accuracies, indicate that image patch-based classifiers truly play an important role in HSIC. As seen from Fig. \ref{fig:SAresult}(e)-(f), there are some misclassified samples in the CapsuleNet and DHCNet results, especially for the grapes untrained class, the Brocoli green weeds 2 class, and the Vinyard untrained class, which demonstrates that there is still improvement space. Moreover, the SSFCN is a more efficient method that can avoid patch extraction and was a new SOTA in 2020. By contrast, the proposed ASPCNet uses a new framework by developing the adaptive spatial unit into the original capsulenet, which can better adapt to the complex spatial characteristics of the HSI. In addition, ASPCNet can obtain the highest classification accuracies in terms of OA, AA, and Kappa with a smaller variance.

\begin{table*}
  \centering
   {
  \caption{Classification Accuracies (\%) on Houston dataset Among Different Methods. Number in Parenthesis Indicates the Standard Variance of the Repeated Experiments. The Best and Second Results Are Shown in Red and Blue colors, Respectively.}
    \begin{tabular}{cc||cccccccc}
    \hline
      \multicolumn{2}{c||}{Classes}  & SVM   & ELM   & EPF   & DeepCNN  & CapsuleNet & DHCNet & SSFCN & ASPCNet \\
    \hline
    \hline
    \multicolumn{1}{c|}{\multirow{15}[2]{*}{CA}} &\cellcolor[rgb]{ 0,  .804,  0} 1     & 96.15(0.69) & 97.70(1.03) & 98.15(1.10) & 97.92(0.75) & 98.11(1.39) & 99.62(0.23) & \textcolor[rgb]{ 0,  0,  1}{\textbf{99.82(0.18)}} & \textcolor[rgb]{ 1,  0,  0}{\textbf{100.00(0.0)}} \\
    \multicolumn{1}{c|}{} &\cellcolor[rgb]{ .498,  .98,  1} 2     & 97.81(0.27) & 98.25(0.75) & 98.30(0.67) & 98.27(0.77) & 98.32(1.51) & \textcolor[rgb]{ 0,  0,  1}{\textbf{99.69(0.20)}} & 99.63(0.18) & \textcolor[rgb]{ 1,  0,  0}{\textbf{99.95(0.05)}} \\
    \multicolumn{1}{c|}{} &\cellcolor[rgb]{ .18,  .545,  .341} \textcolor[rgb]{ 1,  1,  1}{3} & 99.65(0.60) & \textcolor[rgb]{ 1,  0,  0}{\textbf{100.00(0.0)}} & \textcolor[rgb]{ 1,  0,  0}{\textbf{100.00(0.0)}} &  \textcolor[rgb]{ 1,  0,  0}{\textbf{100.00(0.0)}} & 97.86(1.14) & 99.50(0.52) & 99.11(0.57) & \textcolor[rgb]{ 0,  0,  1}{\textbf{99.53(1.31)}} \\
    \multicolumn{1}{c|}{} &\cellcolor[rgb]{ 0,  .545,  0} \textcolor[rgb]{ 1,  1,  1}{4} & 98.47(0.80) & 99.07(0.65) & 98.83(0.96) & \textcolor[rgb]{ 1,  0,  0}{\textbf{99.55(0.23)}} & 95.02(2.10) & 97.64(2.07) & 99.16(0.65) & \textcolor[rgb]{ 0,  0,  1}{\textbf{99.26(0.65)}} \\
    \multicolumn{1}{c|}{} &\cellcolor[rgb]{ .627,  .322,  .176} \textcolor[rgb]{ 1,  1,  1}{5} & 96.33(1.38) & 98.09(0.78) & 97.64(2.09) & 97.29(1.38) & \textcolor[rgb]{ 0,  0,  1}{\textbf{99.82(0.37)}} & 99.41(0.11) & \textcolor[rgb]{ 1,  0,  0}{\textbf{100.00(0.0)}} & 99.45(0.55) \\
    \multicolumn{1}{c|}{} &\cellcolor[rgb]{ 0,  1,  1} 6     & 99.25(0.60) & \textcolor[rgb]{ 1,  0,  0}{\textbf{100.00(0.0)}} & \textcolor[rgb]{ 1,  0,  0}{\textbf{100.00(0.0)}} & \textcolor[rgb]{ 1,  0,  0}{\textbf{100.00(0.0)}} & 96.98(2.70) & 99.26(0.49) & 99.37(0.61) & \textcolor[rgb]{ 0,  0,  1}{\textbf{99.38(0.21)}} \\
    \multicolumn{1}{c|}{} &7     & 91.58(1.91) & 93.75(1.53) & 93.97(2.56) & 94.70(1.12) & 97.24(1.61) & \textcolor[rgb]{ 0,  0,  1}{\textbf{98.57(1.13)}} & 98.47(1.35) & \textcolor[rgb]{ 1,  0,  0}{\textbf{99.55(0.27)}} \\
    \multicolumn{1}{c|}{} &\cellcolor[rgb]{ .847,  .749,  .847} 8     & 89.26(3.30) & 96.15(0.39) & 97.67(1.65) & 96.41(0.28) & \textcolor[rgb]{ 0,  0,  1}{\textbf{97.90(0.57)}} & 96.07(1.58) & 95.20(1.85) & \textcolor[rgb]{ 1,  0,  0}{\textbf{98.84(0.24)}} \\
    \multicolumn{1}{c|}{} &\cellcolor[rgb]{ 1,  0,  0} 9     & 88.32(2.19) & 87.75(2.22) & \textcolor[rgb]{ 0,  0,  1}{\textbf{96.18(1.94)}}  & 91.46(6.08) & 95.70(2.54) & \textcolor[rgb]{ 1,  0,  0}{\textbf{99.45(0.43)}} & \textcolor[rgb]{ 1,  0,  0}{\textbf{99.45(0.55)}} & \textcolor[rgb]{ 1,  0,  0}{\textbf{99.45(0.46)}} \\
    \multicolumn{1}{c|}{} &\cellcolor[rgb]{ .545,  0,  0} \textcolor[rgb]{ 1,  1,  1}{10} & 92.07(1.80) & 86.60(1.58) & 97.02(1.35) & 92.19(4.95) & \textcolor[rgb]{ 1,  0,  0}{\textbf{100.00(0.0)}} & \textcolor[rgb]{ 0,  0,  1}{\textbf{99.24(0.70)}} & \textcolor[rgb]{ 1,  0,  0}{\textbf{100.00(0.0)}} & \textcolor[rgb]{ 1,  0,  0}{\textbf{100.00(0.0)}} \\
    \multicolumn{1}{c|}{} &\cellcolor[rgb]{ 0,  0,  0} \textcolor[rgb]{ 1,  1,  1}{11} & 88.04(1.47) & 89.26(0.75) & 95.05(1.24) & 92.19(2.48) & \textcolor[rgb]{ 1,  0,  0}{\textbf{100.00(0.0)}} & 99.16(0.83) & \textcolor[rgb]{ 0,  0,  1}{\textbf{99.49(0.51)}} & \textcolor[rgb]{ 1,  0,  0}{\textbf{100.00(0.0)}} \\
    \multicolumn{1}{c|}{} &\cellcolor[rgb]{ 1,  1,  0} 12    & 88.00(1.62) & 87.98(1.89) & 88.48(3.13)& 89.93(2.10) & \textcolor[rgb]{ 0,  0,  1}{\textbf{99.74(0.15)}} & 99.26(0.20) & \textcolor[rgb]{ 1,  0,  0}{\textbf{99.77(0.23)}} & 98.27(1.35) \\
    \multicolumn{1}{c|}{} &\cellcolor[rgb]{ .933,  .604,  0} 13    & 74.25(6.36) & 90.47(3.55) & 80.96(4.37) & 87.01(3.26) & \textcolor[rgb]{ 1,  0,  0}{\textbf{100.00(0.0)}} & \textcolor[rgb]{ 0,  0,  1}{\textbf{98.87(0.82)}} & 98.71(0.51) & 97.04(2.44) \\
    \multicolumn{1}{c|}{} &\cellcolor[rgb]{ .333,  .102,  .545} \textcolor[rgb]{ 1,  1,  1}{14} & 96.56(0.43) & 95.72(1.80) & 99.03(1.37) & 96.83(2.60) & \textcolor[rgb]{ 1,  0,  0}{\textbf{100.00(0.0)}} & \textcolor[rgb]{ 0,  0,  1}{\textbf{99.74(0.99)}} & \textcolor[rgb]{ 1,  0,  0}{\textbf{100.00(0.0)}} & \textcolor[rgb]{ 1,  0,  0}{\textbf{100.00(0.0)}} \\
    \multicolumn{1}{c|}{} &\cellcolor[rgb]{ 1,  .498,  .314} 15    & 99.09(0.62) & 99.86(0.20) & \textcolor[rgb]{ 1,  0,  0}{\textbf{100.00(0.0)}} & \textcolor[rgb]{ 0,  0,  1}{\textbf{99.96(0.08)}} & 99.07(0.78) & 99.53(0.14) & \textcolor[rgb]{ 1,  0,  0}{\textbf{100.00(0.0)}} & \textcolor[rgb]{ 1,  0,  0}{\textbf{100.00(0.0)}} \\
    \hline
    \hline
    \multicolumn{2}{c||}{OA}    & 93.01(0.54) & 94.08(0.19) & 96.22(0.33) & 95.35(1.09) & 98.28(0.45) & 99.02(0.23) & \textcolor[rgb]{ 0,  0,  1}{\textbf{99.16(0.12)}} & \textcolor[rgb]{ 1,  0,  0}{\textbf{99.39(0.03)}} \\
    \multicolumn{2}{c||}{AA}    & 92.99(0.71) & 94.71(0.28) & 96.09(0.41) & 95.58(0.79) & 98.38(0.47) & 99.01(0.20) & \textcolor[rgb]{ 0,  0,  1}{\textbf{99.22(0.03)}} & \textcolor[rgb]{ 1,  0,  0}{\textbf{99.31(0.08)}} \\
    \multicolumn{2}{c||}{Kappa}     & 92.43(0.59) & 93.60(0.21) & 95.91(0.35) & 94.97(1.18) & 98.14(0.49) & 99.04(0.24) & \textcolor[rgb]{ 0,  0,  1}{\textbf{99.10(0.13)}} & \textcolor[rgb]{ 1,  0,  0}{\textbf{99.34(0.03)}} \\
    \hline
    \end{tabular}%
  \label{tab:HUresult}}%
\end{table*}%

\begin{figure*}
	\centering{
    \subfigure[]{\includegraphics[scale=0.779]{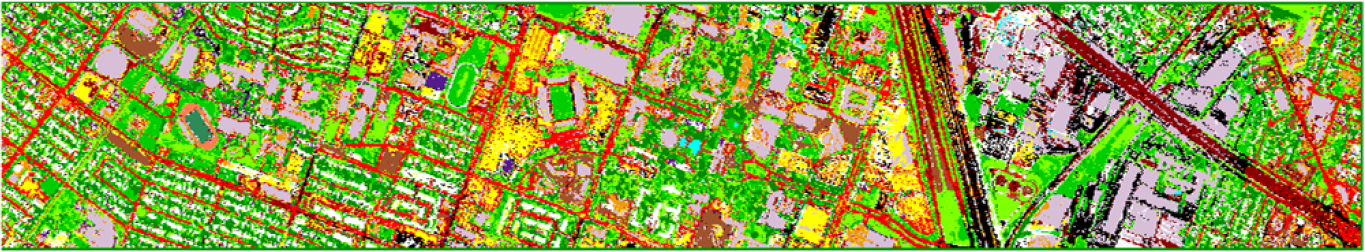}}
    \subfigure[]{\includegraphics[scale=0.779]{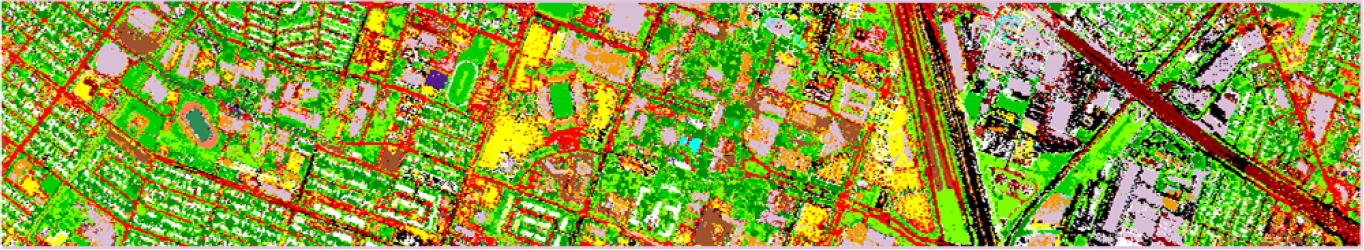}}\par
    \subfigure[]{\includegraphics[scale=0.779]{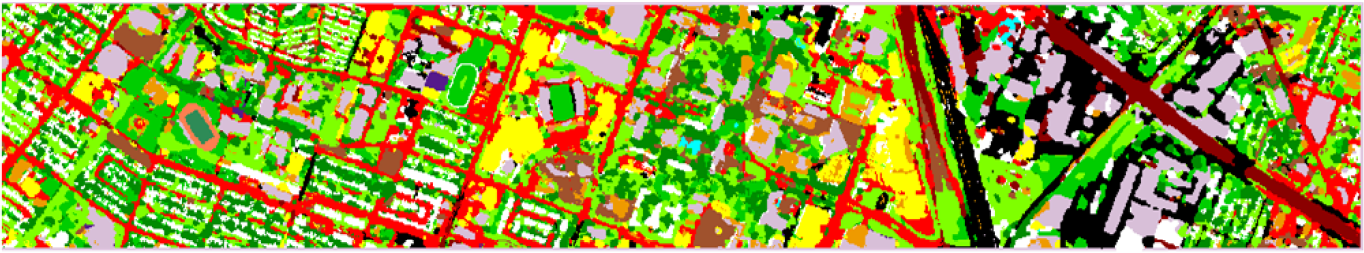}}
    \subfigure[]{\includegraphics[scale=0.779]{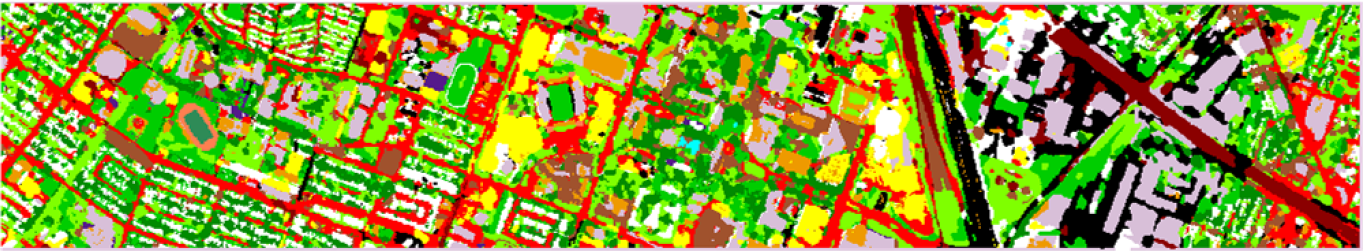}}\par
    \subfigure[]{\includegraphics[scale=0.779]{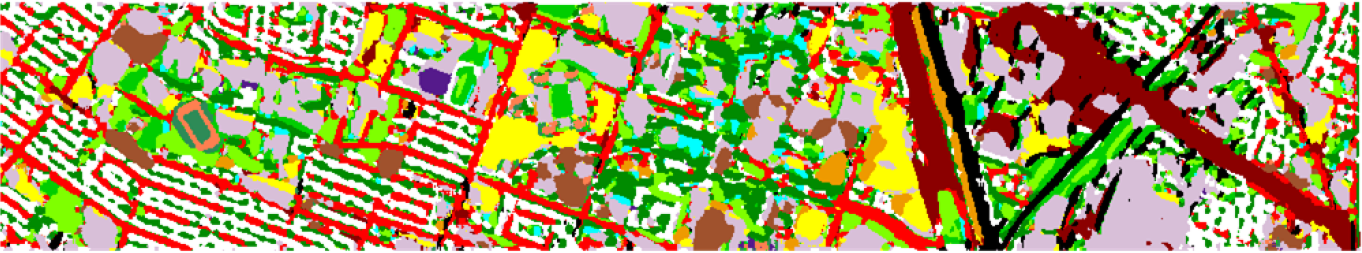}}
    \subfigure[]{\includegraphics[scale=0.779]{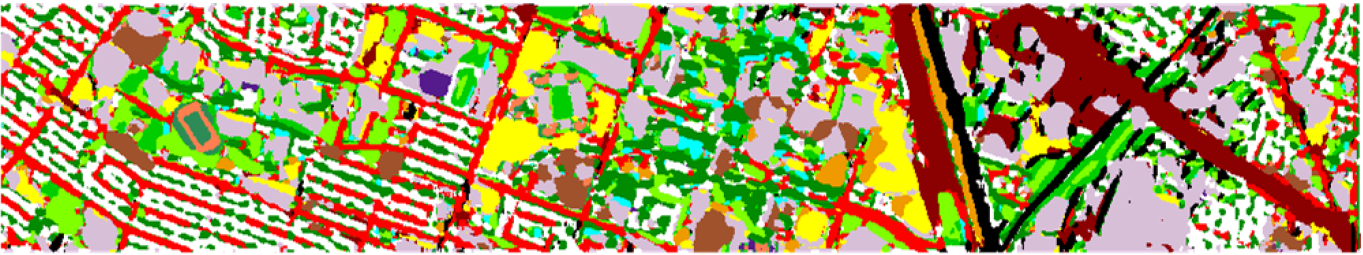}}\par
    \subfigure[]{\includegraphics[scale=0.779]{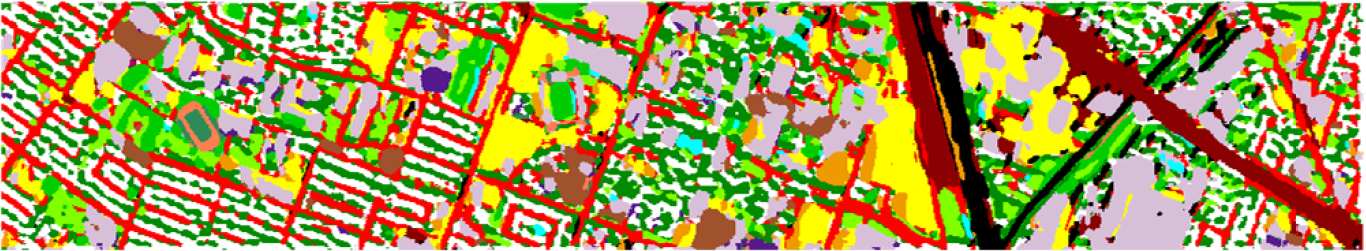}}
    \subfigure[]{\includegraphics[scale=0.779]{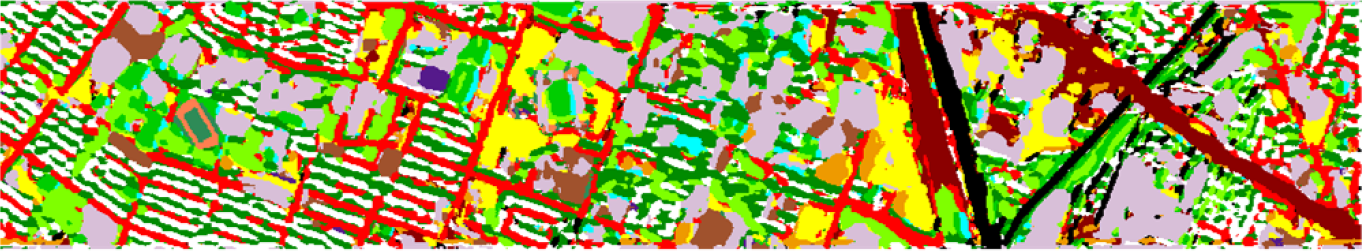}}\par
	\caption{Classification maps obtained by nine methods on the Houston dataset: {(a)} SVM (93.01\%),  {(b)} ELM (94.08\%), {(c)} EPF (96.22\%), {(d)} DeepCNN (95.35\%), {(e)} CapsuleNet (98.28\%), {(f)} DHCNet (99.02\%), {(g)} SSFCN (99.16\%), and {(h)} ASPCNet (99.39\%).}
		\label{fig:HUresult}}
\end{figure*}

% \begin{table*}
%   \centering
%   \caption{The number of Whole Network's Parameters and Memory of Modes for Different Methods.}
%     \begin{tabular}{c|ccccc|ccccc}
%     \hline
%        \multirow{2}[0]{*}{Datasets} & \multicolumn{5}{c|}{Number of Parameters} & \multicolumn{5}{c}{Memory of Modes (MB)} \\
% \cline{2-11}          & DeepCNN  & CapsuleNet & DHCNet & SSFCN & ASPCNet & DeepCNN  & CapsuleNet & DHCNet & SSFCN & ASPCNet \\
%     \hline
%     \hline
%     Salinas    & 3,623,504 & 558,464 & 1,465,588 & 109,458 & 2,885,390 & 41.5  & 6.41  & 16.8  & 1.25  & 32.9\\
%     University of Pavia    & 3,594,665 & 432,384 & 1,463,789 & 96,075 & 1,739,648 & 41.2  & 4.97  & 16.8  & 1.09  & 22.5 \\
%     Houston    & 3,580,175 & 349,056 & 1,465,331 & 101,723 & 2,933,017 & 41.39 & 6.87  & 17.25 & 1.02  & 33.6 \\
%      \hline
%     \end{tabular}
%   \label{tab:netpara}
% \end{table*}

The second experiment is performed on the University of Pavia dataset. First, 200 samples per class are selected randomly as the training samples and the rest are used as test samples (see Table \ref{tab:PUnumber}). Table \ref{tab:PUresult} shows the classification performance of the different methods, and Fig. \ref{fig:PUresult} displays the corresponding classification maps. As seen, there are three classes that can be classified 100\% correctly by the proposed ASPCNet, and ASPCNet achieves the highest accuracies on seven classes, which shows the outstanding performance of ASPCNet. Moreover, as we can see from Fig. \ref{fig:PUresult}, ASPCNet provides more homogeneous classification maps with clear object edges, even for the University of Pavia dataset, with complex structures. 

Then, similar conclusions can be observed on the Houston dataset, for which training samples are shown in Table \ref{tab:HUnumber} and the classification results and maps are displayed in Table \ref{tab:HUresult} and Fig. \ref{fig:HUresult}, respectively. The classification results of ASPCNet are consistently better than those of the compared spectral-spatial classification methods, and the experimental results from the three real datasets indicate the robustness of the proposed ASPCNet method.

Last, experiments conducted on three datasets indicate the effectiveness of the proposed method with a small number of training samples. Here, the number of the training
samples is 40 for each class, which is selected randomly. Table \ref{tab:allsmallsamples} shows the OA, AA, and kappa obtained by different methods of three datasets. As seen from Table \ref{tab:allsmallsamples}, whatever compared with traditional machine learning-based methods or deep learning-based methods, the proposed ASPCNet method can always obtain a better result. More importantly, the proposed method shows obvious
improvements in terms of OA, AA, and kappa with a smaller variance.

\begin{table*}
  \centering
  \caption{Classification Accuracies (\%) of Different Methods on Three Datasets with 40 Training Samples Per Class. Number in Parenthesis Indicates the Standard Variance of the Repeated Experiments. The Best and Second Results Are Shown in Red and Blue colors, Respectively.}
    \begin{tabular}{c|c|cccccccc}
    \hline
    {Datesets} & Index & SVM   & ELM   & EPF   & DeepCNN  & CapsuleNet & DHCNet & SSFCN & ASPCNet \\
    \hline
     \hline
    \multirow{3}[0]{*}{Salinas} & OA    & 85.86(1.29) & 89.89(0.45) & 91.25(2.33) & 92.24(4.39) & 93.00(2.54) & 94.09(0.72) & \textcolor[rgb]{ 0,  0,  1}{\textbf{94.64(2.26)}} & \textcolor[rgb]{ 1,  0,  0}{\textbf{96.97(2.13)}} \\
          & AA    & 90.68(0.40) & 93.59(0.21) & 94.72(0.60) & 94.29(2.15) & 95.61(1.40) & 96.26(0.28) & \textcolor[rgb]{ 0,  0,  1}{\textbf{96.55(1.12)}} & \textcolor[rgb]{ 1,  0,  0}{\textbf{97.73(0.65)}} \\
          & Kappa     & 84.31(1.42) & 88.74(0.50) & 90.28(2.56) & 91.31(4.96) & 92.18(2.85) & 93.42(0.81) & \textcolor[rgb]{ 0,  0,  1}{\textbf{94.02(2.52)}} & \textcolor[rgb]{1,  0,  0}{\textbf{96.81(2.35)}} \\
    \hline
    \hline
    \multicolumn{1}{c|}{\multirow{3}[0]{*}{University of Pavia}} & {OA} & 82.14(1.39) & 84.50(0.57) & 92.88(2.72) & 92.67(2.85) & 93.22(1.97) & 95.29(1.91) & \textcolor[rgb]{ 0,  0,  1}{\textbf{96.73(0.91)}} & \textcolor[rgb]{ 1,  0,  0}{\textbf{97.84(0.65)}} \\
          & AA    & 79.99(0.99) & 81.50(0.78) & 92.39(2.18) & 92.36(1.79) & 92.42(1.58) & 95.30(1.25) & \textcolor[rgb]{ 0,  0,  1}{\textbf{96.66(0.67)}} & \textcolor[rgb]{ 1,  0,  0}{\textbf{97.40(0.52)}} \\
          & Kappa & 77.09(1.67) & 79.83(0.66) & 90.73(3.48) & 90.39(3.67) & 91.07(2.53) & \textcolor[rgb]{ 0,  0,  1}{\textbf{95.23(2.49)}} & 94.86(1.20) & \textcolor[rgb]{ 1,  0,  0}{\textbf{97.13(0.86)}}\\
    \hline
     \hline
    \multirow{3}[0]{*}{Houston} & OA    & 87.47(0.72) & 88.71(0.50) & 91.79(1.10) & 88.98(1.47) & 92.96(1.50) & 93.57(0.54) & \textcolor[rgb]{ 0,  0,  1}{\textbf{94.38(0.58)}} & \textcolor[rgb]{ 1,  0,  0}{\textbf{95.42(0.57)}} \\
          & AA    & 87.27(1.30) & 89.54(0.46) & 91.34(1.69) & 90.90(1.10) & 94.11(1.31) & 94.60(0.63) & \textcolor[rgb]{ 0,  0,  1}{\textbf{95.27(0.44)}} & \textcolor[rgb]{ 1,  0,  0}{\textbf{96.03(0.55)}}\\
          & Kappa & 86.45(0.77) & 87.79(0.53) & 91.12(1.19) & 88.08(1.59) & 92.39(1.62) & 93.04(0.59) & \textcolor[rgb]{ 0,  0,  1}{\textbf{93.93(0.62)}} & \textcolor[rgb]{ 1,  0,  0}{\textbf{95.04(0.61)}}\\
    \hline
    \end{tabular}
  \label{tab:allsmallsamples}
\end{table*}

\begin{table}
  \centering
  {\caption{Classification Accuracies (\%) of Different Steps of The ASPCNet on The University of Pavia dataset With 40 Training Samples Per Class.}
    \begin{tabular}{c|ccc|cc|c}
    \hline
    \multirow{2}[0]{*}{Indexes} & \multicolumn{3}{c|}{ASPCaps-based} & \multicolumn{2}{c|}{ASPConvs-based} & \multirow{2}[0]{*}{ASPCNet} \\
\cline{2-6}          & Dil & Def & Dil-Def & Caps & ConvCaps &  \\
\hline
    OA    & 91.35 &	92.88 &	93.58 &	94.31 &	94.99 &	\textbf{97.81}   \\
    AA    & 88.78 &	92.33 &	93.28 &	95.13 &	95.45 &	\textbf{98.14}   \\
    Kappa     & 88.62 &	90.62 &	91.53 &	94.32 &	95.42 &	\textbf{97.13}   \\
    \hline
    \end{tabular}%
  \label{tab:compared}}%
\end{table}%

% \begin{table*}
%   \centering
%   \caption{Computational Time (in Seconds) of Different Methods, and training samples are shown in Table \ref{tab:SAnumber},\ref{tab:PUnumber} and \ref{tab:HUnumber}.}
%     \begin{tabular}{c|c|cccccccc}
%     \hline
%     Indexes & Datasets & SVM   & ELM   & EPF    & DeepCNN  & CapsuleNet & DHCNet & SSFCN & ASPCNet \\
%     \hline
%     \hline
%     \multirow{3}[0]{*}{Time(s)} & Salinas   & 120.81 & 4.2  & 123.55 & 133.51 & 158.54 & 215.52 & 350.41 & 183.61 \\
%           & University of Pavia    			& 109.81 & 3.34 & 107.13 & 210.18 & 237.89 & 261.34 & 389.45 & 241.92 \\
%           & Houston    						& 230.46 & 6.15 & 238.55 & 323.98 & 381.25 & 270.69 & 623.50 & 313.81 \\
%         \hline
%     \end{tabular}%
%   \label{tab:computingtime}%
% \end{table*}%

\subsection{Comparison of Different Feature Fusion Steps}
To measure the performance of different components of the proposed ASPCNet, an experiment is performed on the University of Pavia dataset with 40 training samples per class. The experimental results are shown in Table \ref{tab:compared}. As seen, Dil, Def, and Dil-Def refer to learning the primary convolution feature using the dilated convolution, deformable convolution, and Dil-Def convolution, respectively. Note that Dil-Def and ASPConv are different. The Dil-Def convolution means that the network conducts dilated convolution first and then deformable convolution later (two layers totally), whereas ASPConv aims to integrate the dilation and deformation operations into one unit (only a single layer). Caps and ConvCaps refer to learning digitalCaps features using the original capsule network and conv-capsule network with digital feature extraction. \par

In the proposed ASPCNet method, we create an ASP unit by taking advantage of both dilated and deformable convolution operations, which can simultaneously expand the receptive field and rotate the sampling location, and effectively avoid grid effects. Note that we develop the ASP unit into a traditional capsulenet during the primarycaps and digitalcaps process. Based on the experimental results observations, the Dil, Def, and Dil-Def operations lead to lower OAs of 91.35\%, 92.88\%, and 93.58\%, respectively. Especially, the classification performance obtained by ASPCNet is better than that of Dil-Def, which indicates that the ASP unit is not a simple combination of the Dil and Def operations, like stacking one to another. By contrast, the ASP unit combines the Dil and Def operations into a unit, improving the classification performance by adaptively enlarging the receptive field and adapting the shapes according to the complex features of HSIs. Based on the ASPConvs modules, we compare the output classifier using traditional caps, ConvCaps, and ASPCaps. According to the experimental results, we find that the proposed ASPCNet can obtain the highest result of the three methods, approximately 97.81\% (traditional caps and ConvCaps can only obtain 94.31\% and 94.99\%, respectively). The results demonstrate that the ASPCaps located in the Caps layer can help improve classification performance.

\begin{table}
  \centering
   \caption{Computational Time (in Seconds) of Different Methods, and training samples are shown in Table \ref{tab:SAnumber},\ref{tab:PUnumber} and \ref{tab:HUnumber}.}
    \begin{tabular}{c|ccc}
    \hline
    \multirow{2}[0]{*}{Methods} & \multicolumn{3}{c}{Datasets} \\
    \cline{2-4}
          & Salinas & University of Pavia & Houston \\
    \hline
    SVM   & 120.81 & 109.81 & 230.46 \\
    ELM   & 4.2   & 3.34  & 6.15 \\
    EPF   & 123.55 & 107.13 & 238.55 \\
    DeepCNN & 133.51 & 210.18 & 323.98 \\
    CapsuleNet & 158.54 & 237.89 & 381.25 \\
    DHCNet & 215.52 & 261.34 & 270.69 \\
    SSFCN & 350.41 & 389.45 & 623.5 \\
    ASPCNet & 183.61 & 241.92 & 313.81 \\
    \hline
    \end{tabular}%
  \label{tab:computingtime}%
\end{table}%
  \subsection{Computing Times of Different Methods}
     The time consumption (in seconds) of the training processes of the proposed ASPCNet and the other compared methods on the three typical hyperspectral datasets are reported in Table \ref{tab:computingtime}. The experimental environments are described at the end of Section \ref{Compared Methods}. The traditional machine learning methods (SVM, ELM, and EPF) and some related deep learning methods (DeepCNN, CapsuleNet, DHCNet, and SSFCN) are compared with the proposed  ASPCNet.  As observed, the proposed ASPCNet method requires more computational time than all the traditional machine learning methods, and the proposed method's main computational cost is due to the dynamic routing algorithm in capsulenet. When compared with DHCNet and SSFCN, the ASPCNet method needs less time on the same benchmark. To make the network more efficient and better able to be transformed for other application fields, searching for a lightweight neural network will be a focus of our future research.

\section{Conclusions}
\label{conclusions}
In this paper, we develop an adaptive spatial pattern capsule network (ASPCNet) for image classification, in which a unique convolutional unit called the ASP unit is used to extract features. On the research basis of the conv-capsulenet, the ASP units are introduced twice into the capsulenet during the classification process. Initially, instead of using a shallow convolutional layer, the proposed ASPCNet uses ASPConvs to extract the relatively high-level features before they are fed into primaycaps, making it easier to transfer hierarchical relationships between low-level and high-level capsules. Furthermore, considering the fixed sampling location of the convolutional kernels, ASPCaps is further introduced to this model, making full use of contextual information adaptively and meeting the requirements of HSIC tasks of complex structures. In addition, the experimental results on three real HSIs demonstrate the superiority of the proposed ASPCNet over several compared methods in terms of the visual qualities of the classification map and quantitative metrics.\par 

{\appendices
\section{Proof of the Section III-C Part}
\label{appendices}
% \subsection{}
\subsection{Hyperparameters for ASPConvs \em{1, 2}} Each block consists of three layers of an ASP layer, a convolution layer, and a rectified linear unit (ReLU) activation function. In Block ASPConv 1, for the ASP layer, the kernel contains 3$\times$3 nonzero weights, 128 filters, stride = 1, and padding = same. For the convolution layer, there are 128 filter kernels of size 1$\times$1,  stride = 2, and padding = same. After the convolution operation, ReLU activation functions are followed. As a practice, a padding operation is conducted to ensure that the feature maps remain the same size before and after convolutions. In ASPConv 2, for the ASP layer, the kernel contains 3$\times$3 nonzero weights, 256 filters, stride = 1, and padding = same. For the convolution layer, the kernel contains 1$\times$1 nonzero weights, 256 filters, stride = 2, and padding = same. After the convolution operation, the ReLU activation functions and batch normalization (BN) layers are followed. 

\subsection{Hyperparameters for ASPCaps \em{1, 2}} There are two ASPC blocks. For ASPCaps 1, the kernel contains 3$\times$3 nonzero weights, 32$\times$4 filters, stride = 1, and padding = same.  After the ASPC operation, ReLU activation functions are followed. For ASPCaps 2, the kernel contains 3$\times$3 nonzero weights, 32$\times$4 filters, stride = 1, and padding = same.}
% % you can choose not to have a title for an appendix
% % if you want by leaving the argument blank
% \section{}

% % use section* for acknowledgment
% \section*{Acknowledgment}

\bibliographystyle{IEEEtran}
\bibliography{refs}

\end{document}